\documentclass[letterpaper, 10 pt, conference]{ieeeconf}  

\IEEEoverridecommandlockouts                              
\usepackage{graphics} 
\usepackage{epsfig} 
\usepackage{times} 
\usepackage{amsmath} 
\usepackage{amssymb}  
\usepackage{dsfont}
\usepackage{tikz}
\usetikzlibrary{arrows, shapes, patterns}
\usepackage{algorithm}
\usepackage{algorithmic}
\usepackage{array,multirow}
\usepackage{booktabs}
\usepackage{pifont}

\title{\LARGE \bf
CEMFormer: Learning to Predict Driver Intentions from In-Cabin and External Cameras via Spatial-Temporal Transformers
}

\author{
Yunsheng Ma$^{1}$,
Wenqian Ye$^{2}$,
Xu Cao$^{2}$,
Amr Abdelraouf$^{3}$,
Kyungtae Han$^{3}$,
Rohit Gupta$^{3}$,
Ziran Wang$^{1}$
\thanks{$^{1}$Y. Ma and Z. Wang are with College of Engineering, Purdue University, West Lafayette, IN. 
{\tt\small \{yunsheng, ziran\}@purdue.edu}}%
\thanks{$^{2}$W. Ye and X. Cao are with Courant Institute of Mathematical Sciences, New York University, New York, NY. 
{\tt\small \{wy2029, irohcao\}@nyu.edu}}%
\thanks{$^{3}$A. Abdelraouf, K. Han, and R. Gupta are with InfoTech Labs, Toyota Motor North America, Mountain View , CA. 
{\tt\small \{amr.abdelraouf, kt, rohit.gupta\}@toyota.com}}%
}%

\usepackage[capitalize]{cleveref}
\crefname{section}{Sec.}{Secs.}
\Crefname{section}{Section}{Sections}
\Crefname{table}{Table}{Tables}
\crefname{table}{Tab.}{Tabs.}

\begin{document}
\maketitle
\thispagestyle{empty}
\pagestyle{empty}

\begin{abstract}
Driver intention prediction seeks to anticipate drivers' actions by analyzing their behaviors with respect to surrounding traffic environments. Existing approaches primarily focus on late-fusion techniques, and neglect the importance of maintaining consistency between predictions and prevailing driving contexts. In this paper, we introduce a new framework called Cross-View Episodic Memory Transformer (CEMFormer), which employs spatio-temporal transformers to learn unified memory representations for an improved driver intention prediction. Specifically, we develop a spatial-temporal encoder to integrate information from both in-cabin and external camera views, along with episodic memory representations to continuously fuse historical data. Furthermore, we propose a novel context-consistency loss that incorporates driving context as an auxiliary supervision signal to improve prediction performance. Comprehensive experiments on the Brain4Cars dataset demonstrate that CEMFormer consistently outperforms existing state-of-the-art methods in driver intention prediction. 
\end{abstract}

\section{Introduction}
Over the past decade, Advanced Driver-Assistance Systems (ADAS) have emerged as an invaluable asset in the automotive industry, significantly enhancing driver safety by seamlessly collaborating with human operators to deliver comprehensive traffic information and timely alerts for hazardous maneuvers. ADAS technologies can aid drivers by identifying potential threats through passive assistance or providing proactive guidance to navigate through safety-critical scenarios by leveraging vehicle-to-everything (V2X) communications~\cite{wang_digital_2020} and augmented reality (AR)~\cite{wang_augmented_2020}.

In response to the growing demand for more advanced safety features, automotive manufacturers have started to develop cutting-edge ADAS capable of anticipating a driver's intentions before they execute a maneuver, thereby preventing accidents. These sophisticated systems rely on a combination of sensors for comprehensive analysis. However, accurately predicting driver intentions remains a formidable challenge, primarily due to several contributing factors.

\begin{itemize}
\item \textit{Complexity of real-world driving scenarios}. 
Numerous factors can hinder a vehicle's ability to perceive its surrounding traffic environment, such as weather, road conditions, lighting, or visibility. Furthermore, traffic situations are dynamic and subject to rapid changes, necessitating continuous monitoring and adaptation.

\item \textit{Constraints of temporal context} 
Driver intention anticipation differs from offline video understanding tasks like action detection, which assume the entire video is accessible during inference. Instead, ADAS must process data causally and in real-time, introducing unique challenges.

\item \textit{Unpredictability of human behavior}.
Human drivers can exhibit unpredictable behavior in a wide range of driving situations, influenced by factors such as distractions, emotional states, inattention, lack of experience, or poor decision-making. This unpredictability can result in hazardous situations on the road, making it difficult for ADAS to accurately anticipate and respond to their actions.
\end{itemize}

Prior research on driver intention anticipation has explored various methods to address these challenges. For instance, following the introduction of the mobility digital twin concept \cite{wang_mobility_2022}, Liao et al.~\cite{liao_driver_2023} further developed a driver digital twin to conduct online prediction of lane-change intentions of drivers in a personalized manner. Wang et al.~\cite{wang_driver_2020} proposed a nonlinear auto-regressive neural network to predict the speed-tracking behaviors of various drivers. Gebert et al.~\cite{gebert_end--end_2019} suggested employing 3D ResNet-101 models to predict driver intentions in an end-to-end fashion. Rong et al.~\cite{rong_driver_2020} introduced a ConvLSTM-based auto-encoder to encode traffic scene motion and fuse features extracted from dual camera inputs using a deep-net classifier.

The aforementioned methods have made significant progress in the field of driver intention anticipation, but they do present certain limitations. Firstly, these approaches mainly focus on combining in-cabin and front-facing view information after processing the data separately during the early stages. This strategy may lead to suboptimal performance, as the individual processing steps might not be specifically designed for optimal integration of information from both sources. Secondly, these methods overlook the importance of maintaining consistency between their predictions and the prevailing driving context. Taking traffic context into account can help alleviate the challenge of reducing uncertainty in predicting driver behaviors. For example, if the vehicle occupies the rightmost lane, the system should not anticipate a lane change to the right.

\begin{figure*}[t]
\centering
\includegraphics[width=0.85\textwidth]{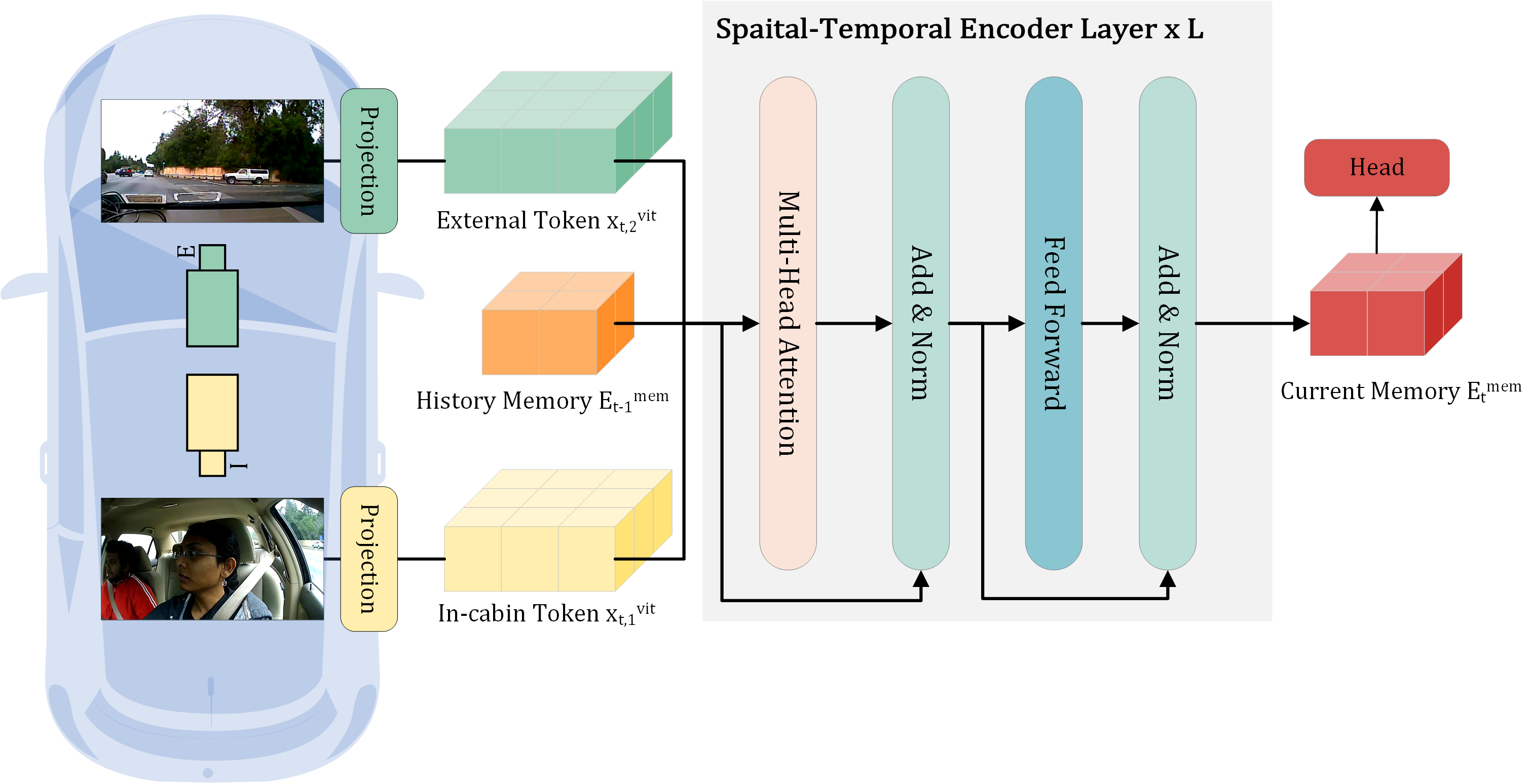}
\caption{\textbf{Overall framework of CEMFormer}. Visualization of the Cross-View Episodic Memory Transformer (CEMFormer) which employs a recurrent architecture to process multi-view camera streams effectively. At each time step $t$, input images are divided into patches, creating a unified sequence. Embeddings from the episodic memory of the previous time step $t-1$ are integrated into this sequence, which serves as input for the spatial-temporal encoder. The output sequence's memory representations are passed to the prediction head and the following time step $t+1$.
}
\label{fig:framework}
\end{figure*}

To address these challenges, we introduce a cross-view episodic memory transformer, named \textbf{CEMFormer}, which effectively aggregates spatio-temporal features from both in-cabin and external cameras as well as historical memory features. The memory representations generated by CEMFormer efficiently support the anticipation of driver intentions. Our CEMFormer model contains three key designs: 
(1) a spatial cross-view encoder that combines spatial features from in-cabin and external camera views, 
(2) an episodic memory module that fuses spatial and temporal information through self-attention mechanisms~\cite{vaswani_attention_2017, cao2023vitasd}, 
and (3) a novel context-consistency loss that utilizes traffic context as supplementary training cues for enhanced prediction accuracy.

Our primary contributions are as follows:
\begin{itemize}
\item We propose CEMFormer, a spatial-temporal transformer encoder that fuses multi-camera and multi-timestamp input into episodic memory representations, addressing the complexities of real-world driving scenarios and constraints of temporal context.

\item We develop a context-consistent loss that enhances the model's ability to employ traffic context as an auxiliary supervision signal during training, reducing uncertainty in predicting driver intentions.

\item We assess the proposed CEMFormer on the Brain4Cars benchmark. Our CEMFormer consistently achieves superior performance compared to previous state-of-the-art methods. For instance, CEMFormer attains $87.09\%$ F1 score with approximately $60\%$ fewer parameters, outperforming the previous best method by 2.8 points. Furthermore, the lightweight model architecture allows for an inference speed of 15 FPS, making it suitable for real-time deployment.
\end{itemize}

\newcommand{\cmark}{\ding{51}}%
\newcommand{\xmark}{\ding{55}}%
\newcommand{\x}{\mathbf{x}}
\newcommand{\yhat}{\hat{y}_t}
\newcommand{\z}{\mathbf{z}}
\newcommand{\aaa}{\mathbf{a}}
\newcommand{\cc}{\mathbf{c}}
\newcommand{\p}{\mathbf{p}}
\newcommand{\ttt}{\mathbf{t}}
\newcommand{\mem}{\text{mem}}
\newcommand{\X}{\mathcal{X}}
\newcommand{\Y}{\mathcal{Y}}
\newcommand{\C}{\mathcal{C}}
\newcommand{\SSS}{\mathcal{S}}
\newcommand{\A}{\mathcal{A}}

\section{Methodology}
\subsection{Preliminaries}
Given multi-view streaming camera inputs and traffic-related context information, our objective is to predict a future event based on observations up to the present time. These future events fall into one of several predefined categories. We denote the input space as $\X$, the output space as $\Y$, and the context space as $\C$. During training, a set of $N$ training samples $\left\{(\x_1,\x_2,...,\x_{T_j})j,y_j,\cc_j\right\}_{j=1}^N$ is provided, where $\x_t\in\X$ represents the observation at time $t$. $y\in\Y$ is the ground-truth label of the event that occurs at the end of the video at time $T_j$. $\cc\in\C$ is the vector containing auxiliary context information, which is utilized during training. During inference, however, the algorithm processes each incoming video frame in an online manner. Specifically, the online prediction system receives $\x_t$ at each time step, with the goal of predicting the event $y$ that will occur at time $T$ given only past and current observations $(\x_1,\x_2,...,\x_t)$, where $t<T$.

We propose a novel transformer-based framework for online driver intention anticipation, designed to effectively aggregate spatio-temporal features from in-cabin and external cameras as well as episodic memory representations using attention mechanisms. As illustrated in~\cref{fig:framework}, CEMFormer comprises $L$ encoder layers, each adopting the conventional structure from vision transformers~\cite{dosovitskiy_image_2021} with some tailored designs. 

\newcommand{\sizecci}{0.24}
\newcommand{\sizeccj}{0.8}

\begin{figure*}[t]%
\centering
\begin{minipage}{\textwidth}
\centering
\begin{minipage}{\sizecci\textwidth}
    \centering
    \scriptsize
    \includegraphics[width=\sizeccj\linewidth]{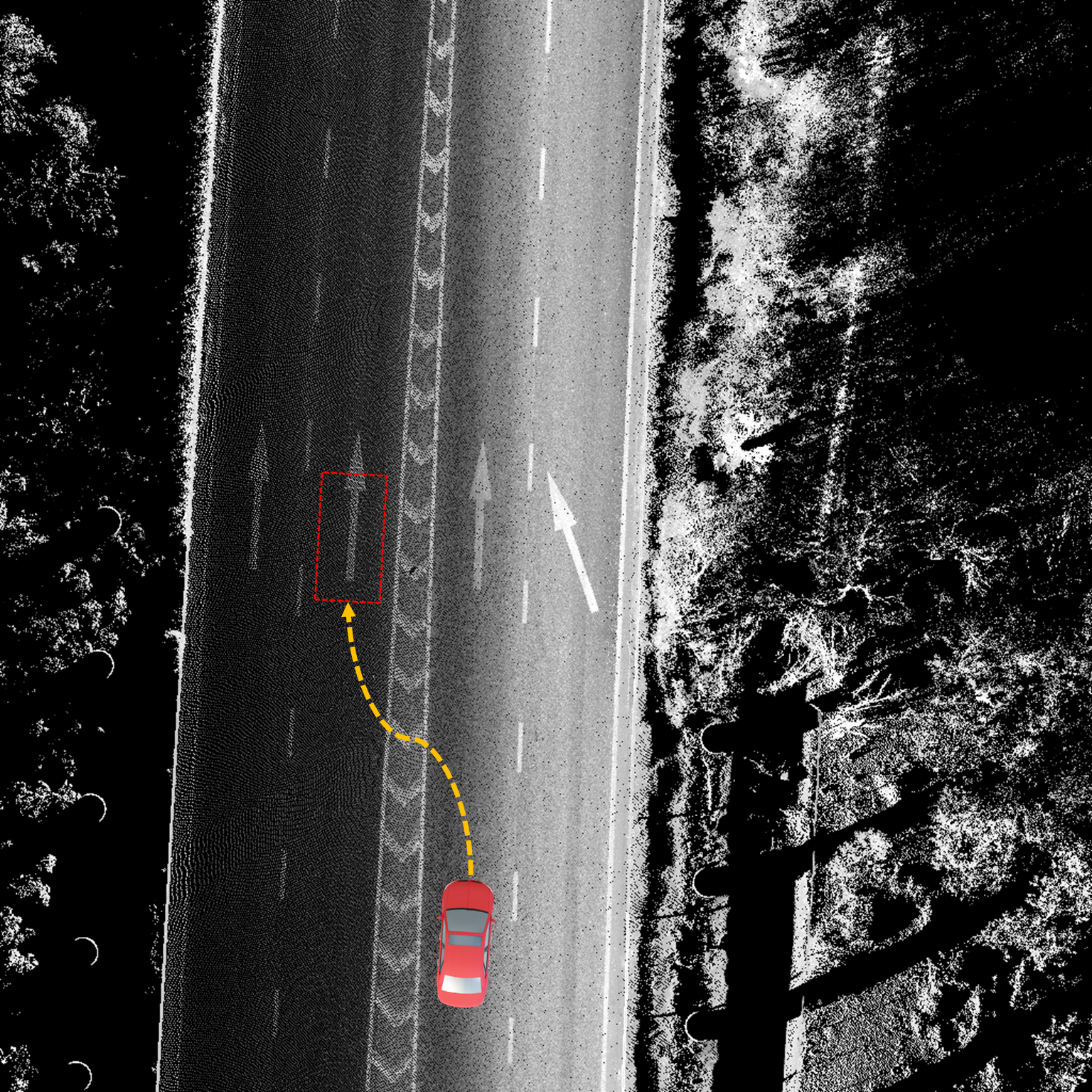}\\
    \textit{left lane change} $\land$ \\
    \textit{left most lane}
\end{minipage}
\begin{minipage}{\sizecci\textwidth}
    \centering
    \scriptsize
    \includegraphics[width=\sizeccj\linewidth]{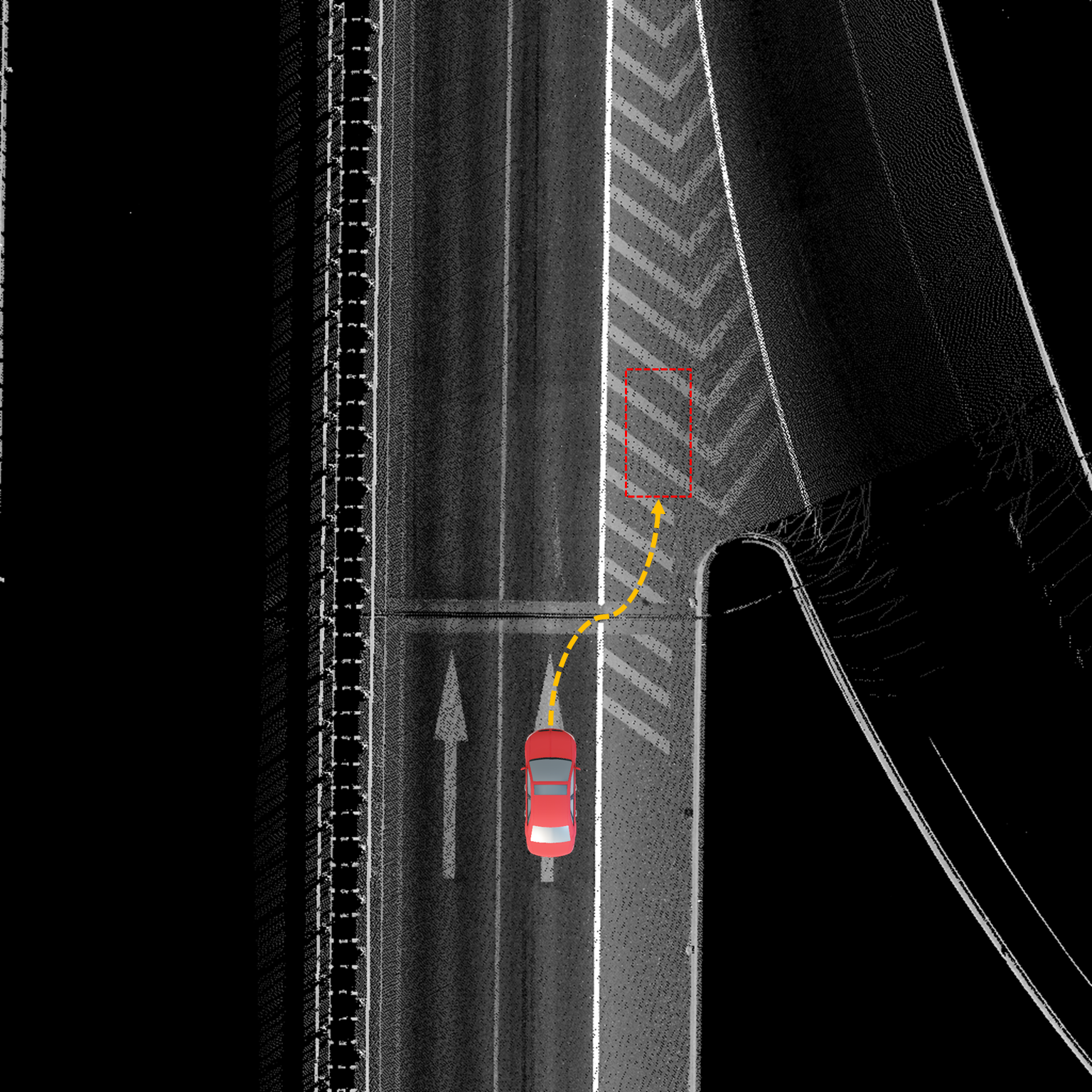}\\
    \textit{right lane change} $\land$ \\
    \textit{right most lane}
\end{minipage}
\begin{minipage}{\sizecci\textwidth}
    \centering
    \scriptsize
    \includegraphics[width=\sizeccj\linewidth]{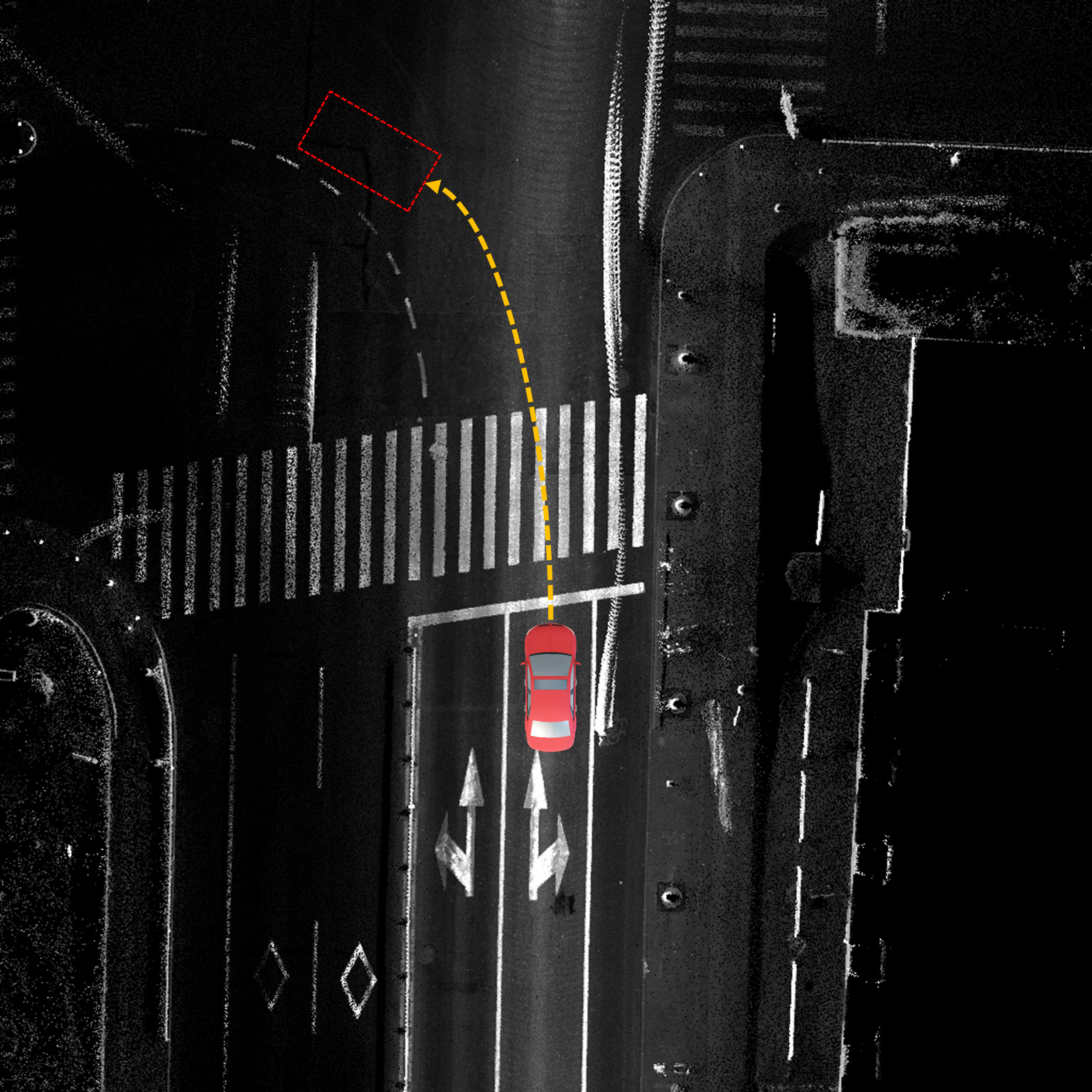}\\
    \textit{left turn} $\land$ \\
    ($\lnot$ \textit{left most lane}) $\land$ \textit{right most lane}
\end{minipage} 
\begin{minipage}{\sizecci\textwidth}
    \centering
    \scriptsize
    \includegraphics[width=\sizeccj\linewidth]{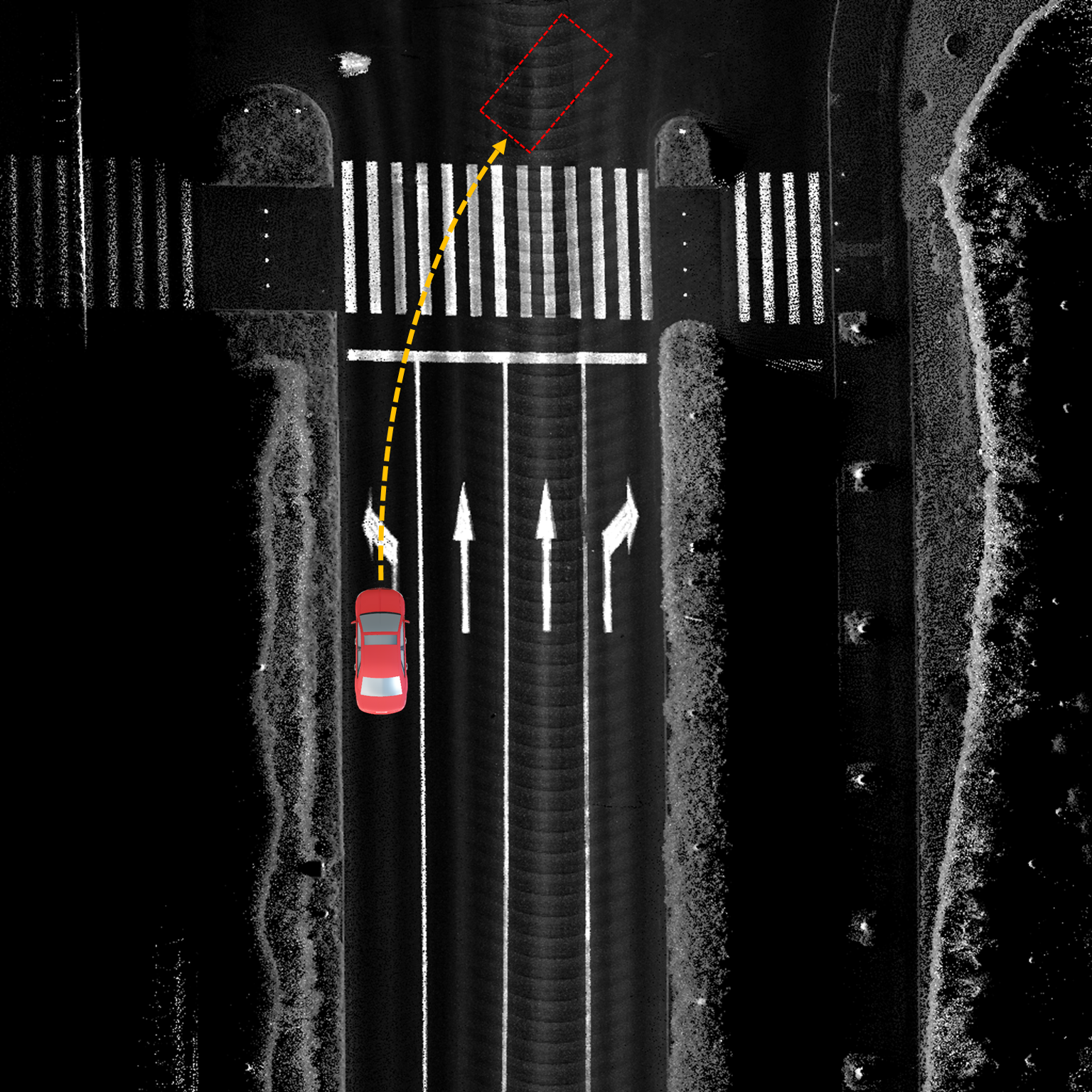}\\
    \textit{right turn} $\land$ \\
    ($\lnot$ \textit{right most lane}) $\land$ \textit{left most lane}
\end{minipage}
\end{minipage} 
\caption{
Visualization of several scenarios where a predicted maneuver (e.g., left lane change) conflicts with the current traffic context (e.g., being in the leftmost lane), leading to increased penalties in the context consistency loss.}
\label{fig:cc_loss}
\end{figure*}

\subsection{Multi-View Embeddings}
Firstly, distinct patch projection layers are applied to each input view. Suppose we have the observation $\x_t$ at time $t$, which contains $M$ views:
\begin{equation}
\x_t=\left\{\x_{t,m}\in\mathbb{R}^{C\times H_m\times W_m}\right\}_{m=1}^M,
\end{equation}
where $H_m$ and $W_m$ represent the height and width of the input image from view $m$, and $C$ denotes the number of channels. For simplicity, we omit $t$ in this paragraph. The image $\x_m$ is then divided into a grid of $N_m$ patches, each with a size of $P^2\cdot C$, where $(P,P)$ is the patch size and $N_m=H_mW_m/P^2$ is the number of patches. The patches are subsequently flattened and linearly projected to $D$-dimensional tokens:
\begin{align}
\x_m^{\textit{vit}}&=[\x_m^1E_m,...,\x_m^{N_m}E_m]+E_m^{\textit{pos}}&m=1,2,...,M,
\end{align}
where $E_m\in\mathbb{R}^{(P^2\cdot C)\times D}$ is the projection matrix, and $E_m^{\textit{pos}}\in\mathbb{R}^{N_m\times D}$ represents the position embedding. Inputs from multiple views are then concatenated into a combined sequence $\z^{0}$:
\begin{align}
\z^{0}&=[\x_1^{\textit{vit}};\x_2^{\textit{vit}};\cdots;\x_M^{\textit{vit}}],
\end{align}
which is subsequently fed into the spatial-temporal encoder.

\subsection{Episodic Memory}
Episodic memory pertains to the capacity to recall latent representations from the past time series~\cite{sandler_fine-tuning_2022}. CEMFormer incorporates this concept to process camera inputs online, allowing it to retain and reference prior context while facilitating information flow between iterations. Consequently, CEMFormer can make predictions based on both previous and current frames, resulting in enhanced robustness and accuracy.

Specifically, $K$ episodic memory embeddings $E_t^\mem\in\mathbb{R}^{K\times D}$ are prepended to the input sequence:
\begin{equation}
\bar{\z}_t^{0}=[E_t^\mem;\z_t^{0}],
\end{equation}
The spatial-temporal encoder then aggregates the features from in-cabin and external cameras as well as episodic memory representations by stacking $L$ transformer blocks:
\begin{equation}
\z_t^{L}=\operatorname{SpatialTemporalEncoder}\left(\bar{\z_t}^{0}\right)
\end{equation}
In the last layer, to ensure information flow between frames, the current episodic memory representations are passed to the input sequence of the next moment:
\begin{equation}
E_{t+1}^\mem=\z_{t,1:K}^{L},\bar{\z}_{t+1}^{0}=[E_{t+1}^\mem;\z_{t+1}^{0}].
\end{equation}

\subsection{Context-Consistency Loss}
The episodic memory representations at the output of the spatial-temporal encoder are fed into a prediction head to generate the final outputs. 
Directly optimizing the standard cross-entropy loss $\ell^{ce}$ can lead to incorrect predictions of some scenarios into categories that conflict with the traffic context. To address this issue, we propose a new context-consistency (CC) loss, which applies a penalty for making such wrong predictions. Specifically, let $\mathcal{S}=\left\lbrace(r,\A) \mid r\in\Y,\A\subset\C\right\rbrace$ be the set of contradicting scenarios, which is a subset of false positive cases. The CC loss can be defined as:
\begin{equation}
{\ell^{cc}} = 
-{\sum\limits_{(r,\A)\in\mathcal{S}}}
{\mathds{1}_{[\cc\in\A]}\log (1-p_r)},
\label{eq:cc_loss}
\end{equation}
where $\cc$ is the current traffic context, $\mathds{1}_{[\cc\in\A]}$ is a binary indicator, and $p_r$ is the predicted probability of event $r$ given by the model. 

Following~\cite{jain_recurrent_2016}, we take advantage of the exponentially growing loss to encourage the model to predict early while ensuring that it does not over-fit the training data when there is insufficient context for anticipation. Combining the cross-entropy loss, the context-consistent loss, and the exponentially growing loss, we refer to the unified loss function as the joint Context-Consistent cross entropy loss:
\begin{equation}
\mathcal{L}_{\text{joint}}={\sum\limits_{i = 1}^N}{\sum\limits_{t = 1}^T}{e^{-(T-t)}(\ell_i^{cc} + \ell_i^{ce})}.
\label{eq:ccce_loss}
\end{equation}

Since the derivatives with respect to all parameters can be computed, we can effectively train the proposed CEMFormer using an off-the-shelf optimizer to minimize the loss function with back-propagation through time (BPTT) as in~\cite{hochreiter_long_1997}.

\section{Experiments and Results}
\begin{table*}[!tb]
\centering
\caption{Comparison of the proposed CEMFormer with various state-of-the-art (SOTA) methods. The top-performing method for each setting is highlighted in \textbf{bold}. $\downarrow$: Lower values are better. $\uparrow$: Higher values are better. $^*$: Results obtained from the original papers. Results for the five-fold evaluation are presented as "AVG $\pm$ SD".}
\begin{tabular}{l|c|ccc}
    \toprule
    Data Source & Method & Param.(M) $(\downarrow)$ & Accuracy $(\uparrow)$ & $F1$ $(\uparrow)$\\
    \midrule
    \multirow{3}{*}{in-cabin only} 
    &Gebert et al.~\cite{gebert_end--end_2019}$^*$ &240.26 & 0.8310 $\pm$ 0.0250 & 0.8170 $\pm$ 0.0260\\
    &Rong et al.~\cite{rong_driver_2020}$^*$ & \textbf{46.22} & 0.7740 $\pm$ 0.0002 & 0.7549 $\pm$ 0.0002\\
    &\textbf{CEMFormer (ours)} & 86.6 & \textbf{0.8447} $\pm$ \textbf{0.0598} & \textbf{0.8266} $\pm$ \textbf{0.0540}\\
    \midrule
    \multirow{3}{*}{external only} 
    &Gebert et al.~\cite{gebert_end--end_2019}$^*$ &240.26 & 0.5320 $\pm$ 0.0500 & 0.4340 $\pm$ 0.0900\\
    &Rong et al.~\cite{rong_driver_2020}$^*$ & 160.41 & 0.6087 $\pm$ 0.0001 & \textbf{0.6638 $\pm$ 0.0003}\\
    &\textbf{CEMFormer (ours)} & \textbf{86.6} & \textbf{0.6475} $\pm$ \textbf{0.0282} & 0.6631 $\pm$ 0.0219 \\
    \midrule
    \multirow{3}{*}{in-cabin $\&$ external} 
    &Gebert et al.~\cite{gebert_end--end_2019}$^*$ & 325.52 & 0.7550 $\pm$ 0.0240 & 0.7320 $\pm$ 0.0220\\
    &Rong et al.~\cite{rong_driver_2020}$^*$ & 212.92 & 0.8398 $\pm$ 0.0001 & 0.8430 $\pm$ 0.0001\\
    &\textbf{CEMFormer (ours)} & \textbf{87.3} & \textbf{0.8537} $\pm$ \textbf{0.0295} & \textbf{0.8709} $\pm$ \textbf{0.0023}\\
    \bottomrule
\end{tabular}
\label{tab:sota}
\end{table*}

\subsection{Dataset and Setup}
We assess the performance of our proposed method for maneuver anticipation using the publicly available Brain4Cars dataset~\cite{jain_brain4cars_2016}. This dataset comprises 594 video clips, showcasing both in-cabin and forward-facing views of a vehicle\footnote{Though it was reported that the dataset includes 700 videos, a portion of them are missing in the public release.}. The dataset encompasses five driver maneuver categories, which defines the output space for our experiments $\Y=$\{\textit{go straight, left lane change, left turn, right lane change, right turn}\}. Lane changes and turns are annotated with the maneuver's start time, corresponding to when the wheel touches the lane marking or when the vehicle begins to yaw at the intersection, respectively~\cite{jain_recurrent_2016}.

Based on the available traffic context information in the dataset, we define the traffic context vector $\cc\in\C\subseteq\mathbb{R}^3$, as a three-dimensional binary vector. Each dimension represents whether the ego vehicle is \textit{in the left-most lane, in the right-most lane}, and \textit{near an intersection}, respectively. Additionally, the set $\SSS$ of contradicting scenarios in~\cref{eq:cc_loss} is formally defined as $\SSS=\{$(\textit{left lane change}, $(1,\cdot,\cdot)$),
(\textit{right lane change}, $(\cdot,1,\cdot)$),
(\textit{left turn}, $(0,1,\cdot)$),
(\textit{right turn}, $(1,0,\cdot)$),
(\textit{left turn}, $(\cdot,\cdot,0)$),
(\textit{right turn}, $(\cdot,\cdot,0)$)$\}$, and is visualized in ~\cref{fig:cc_loss}.

To ensure the reliability of the results, we employ a 5-fold cross-validation in all our experiments, which is consistent with previous studies. The final evaluation metrics include the average accuracy and F1 score, along with their standard deviations.


\begin{table}[!tb]
\centering
\caption{Ablation study comparing the performance of CEMFormer with and without the inclusion of episodic memory (EM) and context consistency (CC).}
\begin{tabular}{cc|cc}
    \toprule
    \multicolumn{2}{c}{Module} \vrule & \multirow{2}{*}{Accuracy $(\uparrow)$}   & \multirow{2}{*}{$F1$ $(\uparrow)$}\\
    \cmidrule{1-2}
    EM & CC &\\
    \midrule
    \xmark & \xmark & 0.7640 $\pm$ 0.0059 & 0.7599 $\pm$ 0.0161\\
    \xmark & \cmark & 0.7751 $\pm$ 0.0424 & 0.8041 $\pm$ 0.0296\\
    \cmark & \xmark & 0.8176 $\pm$ 0.0051 & 0.8143 $\pm$ 0.0262\\
    \cmark & \cmark & \textbf{0.8537} $\pm$ \textbf{0.0295} & \textbf{0.8709} $\pm$ \textbf{0.0023}\\
    \bottomrule
\end{tabular}
\label{tab:ab_m}
\end{table}

\begin{table}[!tb]
\centering
\caption{Ablation study examining the impact of varying the number of episodic memory tokens ($K$) on CEMFormer's performance.}
\begin{tabular}{c|cc} 
\toprule
K & Accuracy $(\uparrow)$ & $F1$ $(\uparrow)$\\
\midrule
2 & 0.8304 $\pm$ 0.0187 & 0.8623 $\pm$ 0.0010 \\
4 & \textbf{0.8537} $\pm$ \textbf{0.0295} & \textbf{0.8709} $\pm$ \textbf{0.0023}\\
8 & 0.8511 $\pm$ 0.0473 & 0.8681 $\pm$ 0.0172\\
\bottomrule
\end{tabular}
\label{tab:abk}
\end{table}

\subsection{Implementation Details}
For our experiments, we initialize our model using the DINO ViT-B/16~\cite{caron_emerging_2021} pre-trained weighs, which was trained on ImageNet~\cite{deng_imagenet_2009}. We adopt AdamW \cite{loshchilov_decoupled_2019} as the optimizer with a weight decay of 0.05 and apply a cosine learning rate scheduler~\cite{ilya_loshchilov_sgdr_2017} with the base learning rate set to $5\times 10^{-5}$. Both the in-cabin and external vehicle camera streams have a resolution of $224\times 224$. With a patch size of $16\times 16$, this results in a total of 392 patches. We empirically set the number of memory tokens $K=4$ (Ablation study is in \cref{sec:ab}). For data augmentation, we divide each video into $T$ segments of equal duration and randomly sample one frame from each segment, where $T$ is the video length in seconds, drawing inspiration from \cite{wang_temporal_2016}. As a frame-level data augmentation strategy, we also employ simple random crop with random horizontal flip, introduced in \cite{touvron_deit_2022}. Owing to the limited size of the Brain4Cars dataset, we only fine-tune the multi-head self-attention layers in the spatial-temporal encoder, as recommended in \cite{touvron_three_2022}. Our model is trained for 200 epochs on a single NVIDIA RTX 3090 Ti GPU with a batch size of 10.

\subsection{Comparison with State-of-the-Art}
~\cref{tab:sota} presents the comparison results on the Brain4Cars test set. We compare CEMFormer to two other widely used end-to-end methods~\cite{gebert_end--end_2019,rong_driver_2020}, as they have outperformed traditional machine learning approaches in driver intention prediction tasks\footnote{For a fair comparison, we do not consider results from~\cite{jain_recurrent_2016} as part of its training data is not accessible.}. CEMFormer achieves the highest accuracy of 85.37\% and an F1 score of 87.09\% with the multi-view inputs. Furthermore, we observe that CEMFormer surpasses the other two methods in terms of the number of parameters. These results indicate that the proper use of an episodic memory-guided architecture allows the model to learn complex spatial-temporal relationships from both in-cabin and external driving views. This is in contrast to previous work such as \cite{gebert_end--end_2019}, which claimed that outside views were not helpful for driver intention prediction tasks. Additionally, our model significantly reduces the number of parameters compared to previous methods while maintaining a compact size, even when incorporating additional views. This demonstrates the scalability of the proposed model.

\newcommand{\vspacefigtwo}{\vspace{0.5mm}}
\newcommand{\fontsizefigtwo}{\scriptsize}
\newcommand{\sizeafigtwo}{0.19}
\newcommand{\sizerfigtwo}{0.5}
\newcommand{\rech}{1.8}
\newcommand{\recw}{0.96\linewidth}
\newcommand{\recl}{0}
\newcommand{\lineend}{0.75\textwidth}

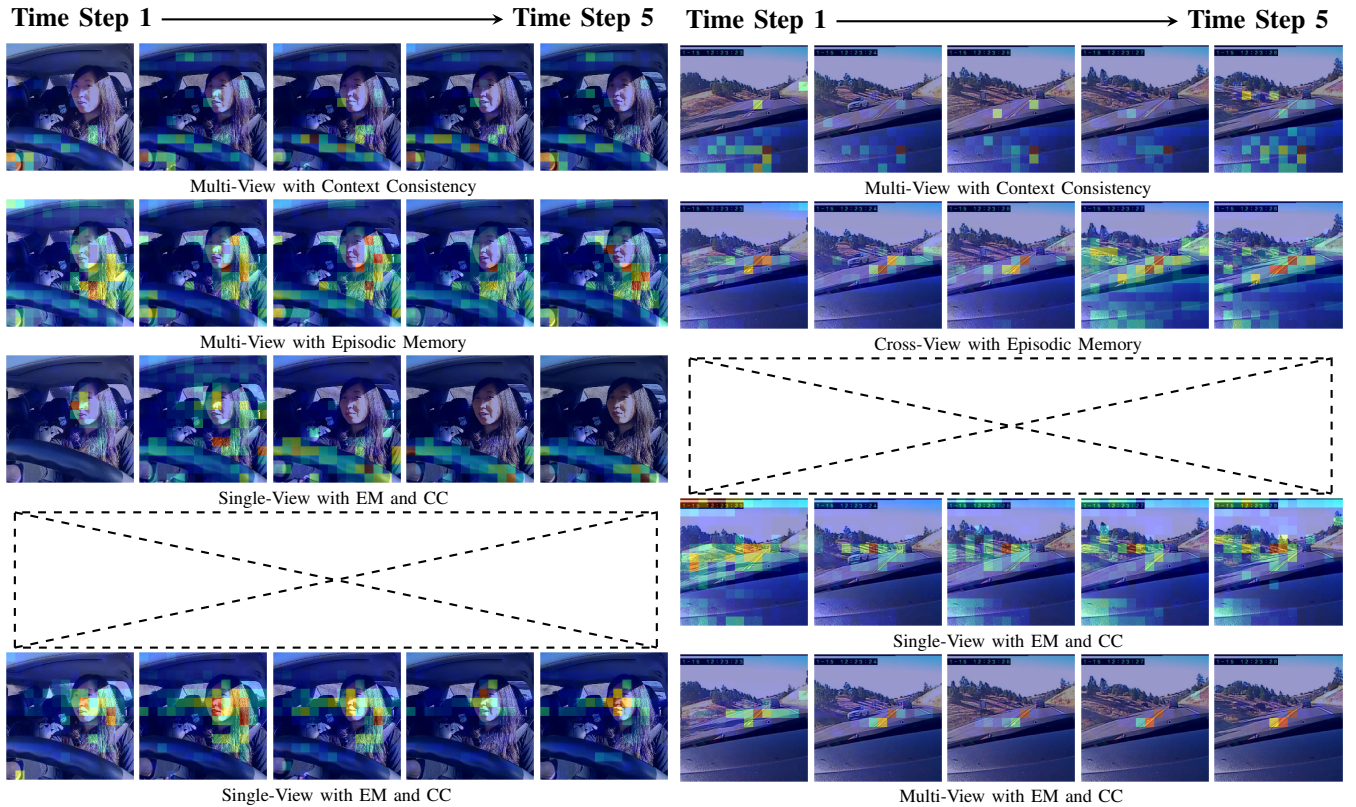
\begin{figure*}[!tb]%
\begin{minipage}{\textwidth}
\begin{minipage}{\sizerfigtwo\textwidth}
\centering
\begin{minipage}{\textwidth}
\begin{tikzpicture}[
    arrow/.style = {thick,-stealth}]
\node (A) at (0, 0) {\textbf{Time Step 1}};
\node (B) at (\lineend, 0) {\textbf{Time Step 5}};
\draw [arrow] (A) -- (B);
\end{tikzpicture}
\vspacefigtwo
\end{minipage}
\begin{minipage}{\textwidth}
\centering
\fontsizefigtwo
\includegraphics[width=\sizeafigtwo\linewidth,height=\sizeafigtwo\linewidth]{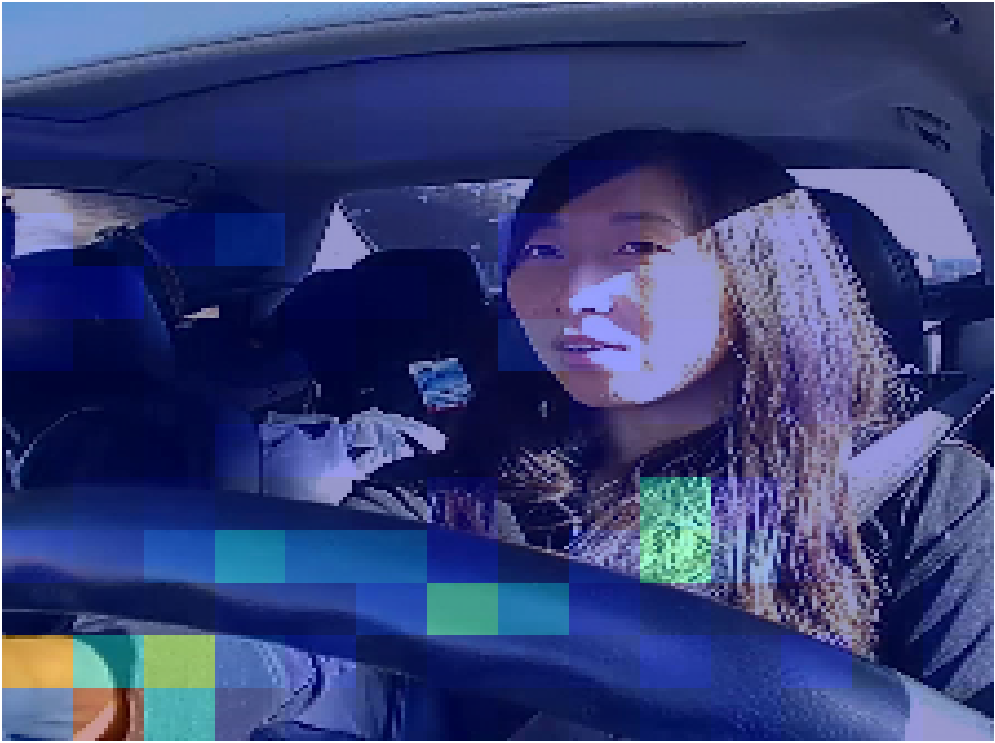}
\includegraphics[width=\sizeafigtwo\linewidth,height=\sizeafigtwo\linewidth]{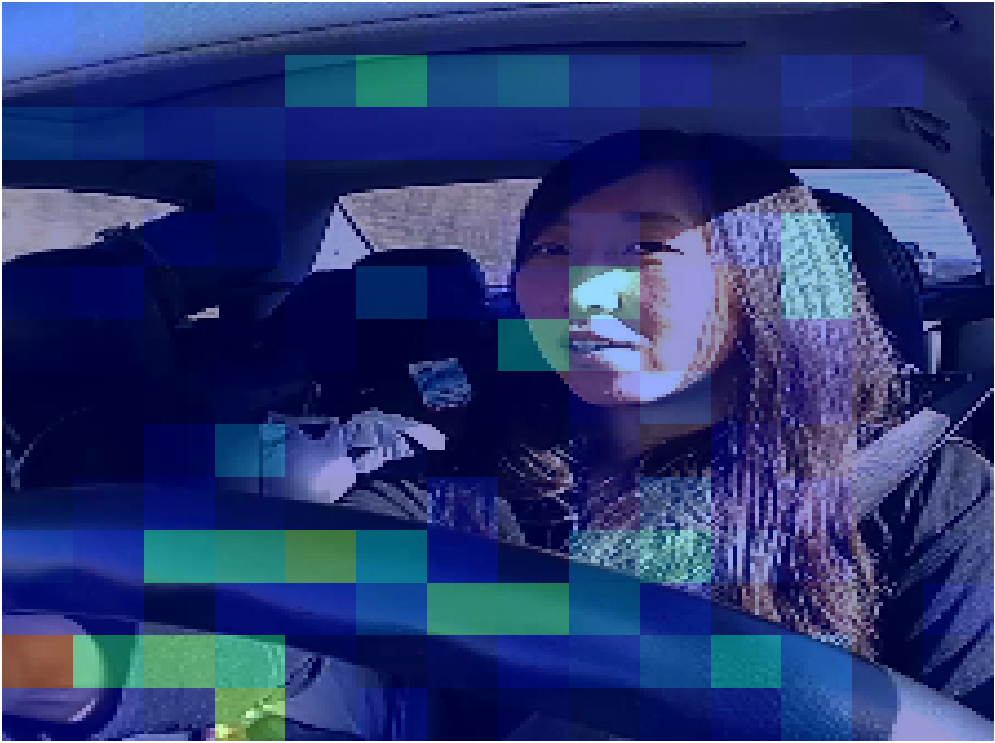}
\includegraphics[width=\sizeafigtwo\linewidth,height=\sizeafigtwo\linewidth]{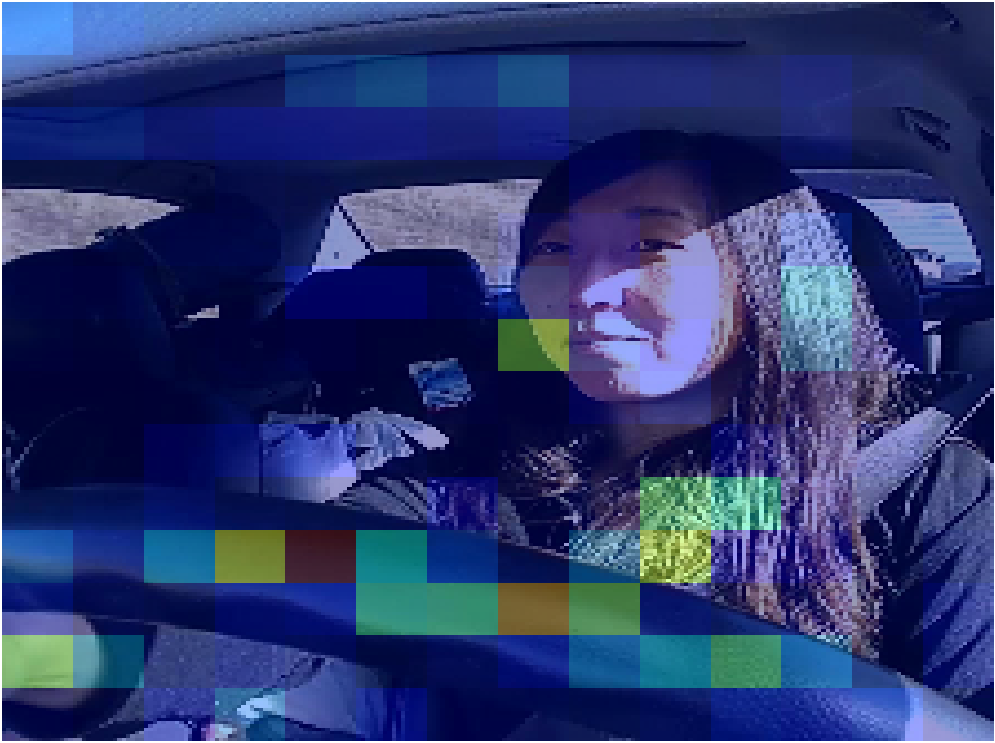}
\includegraphics[width=\sizeafigtwo\linewidth,height=\sizeafigtwo\linewidth]{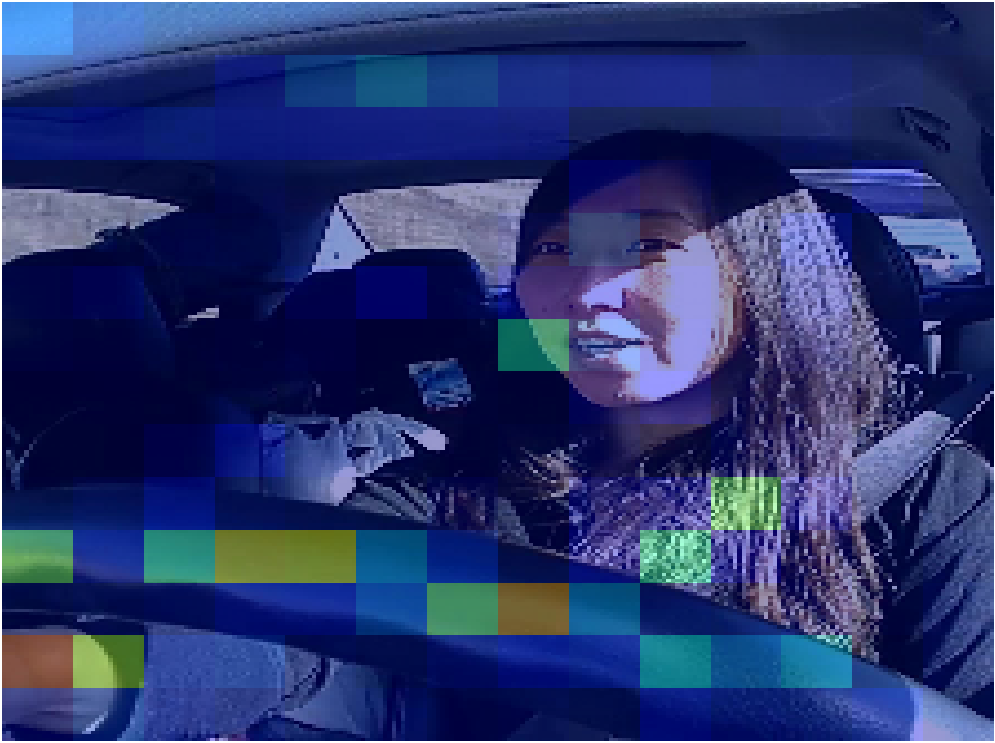}
\includegraphics[width=\sizeafigtwo\linewidth,height=\sizeafigtwo\linewidth]{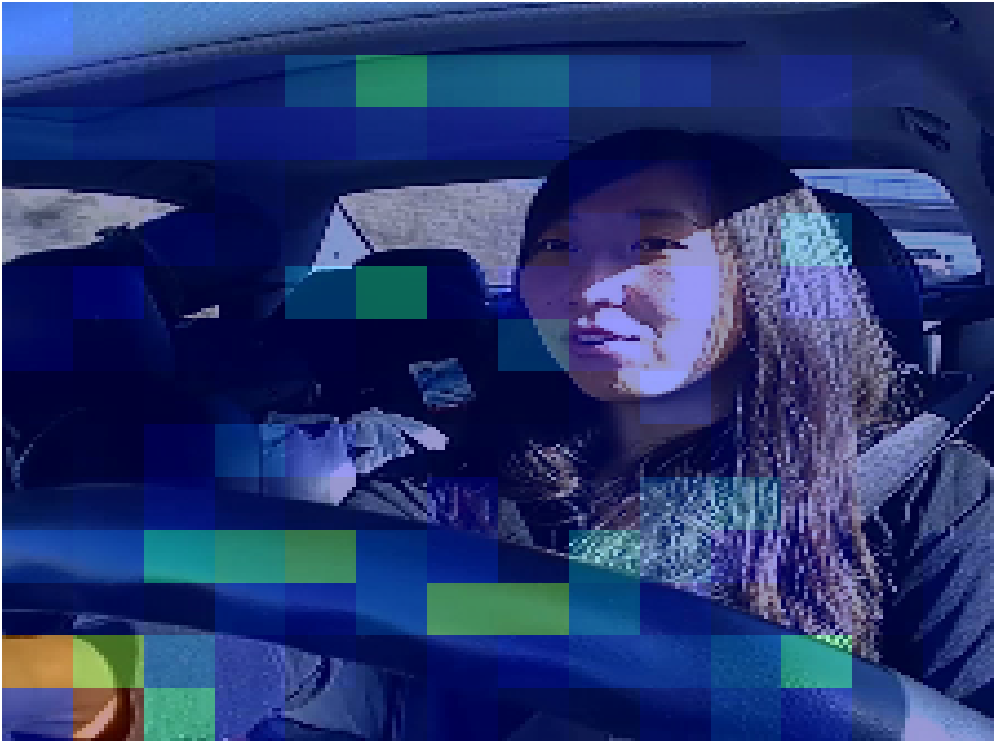}
\\
Multi-View with Context Consistency
\vspacefigtwo
\end{minipage}
\begin{minipage}{\textwidth}
\centering
\fontsizefigtwo
\includegraphics[width=\sizeafigtwo\linewidth,height=\sizeafigtwo\linewidth]{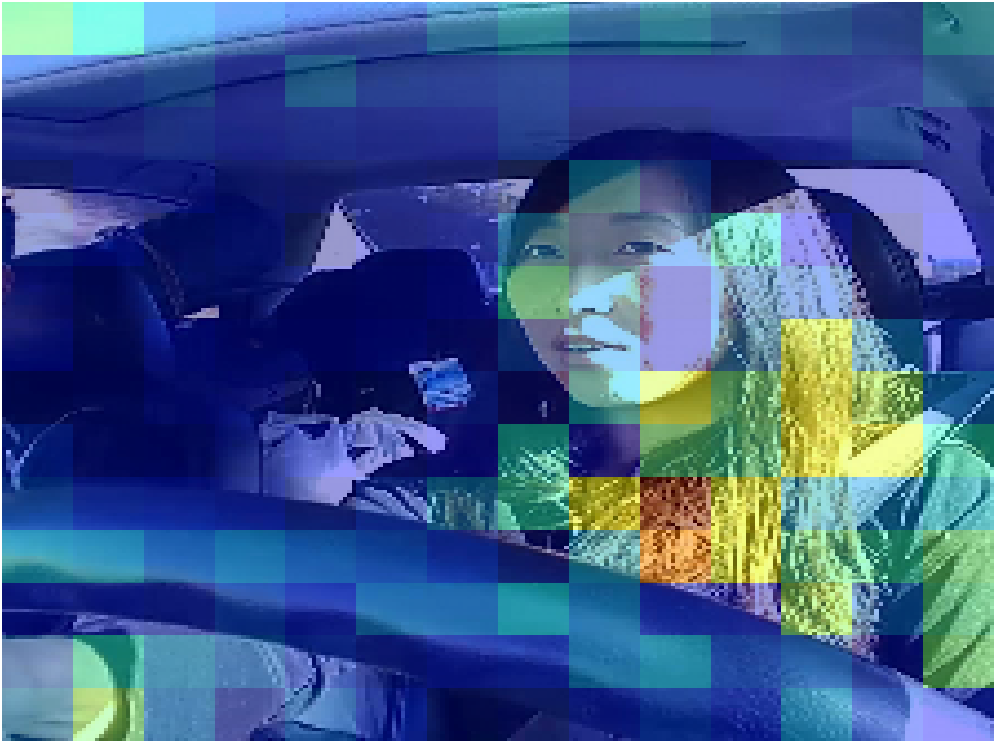} 
\includegraphics[width=\sizeafigtwo\linewidth,height=\sizeafigtwo\linewidth]{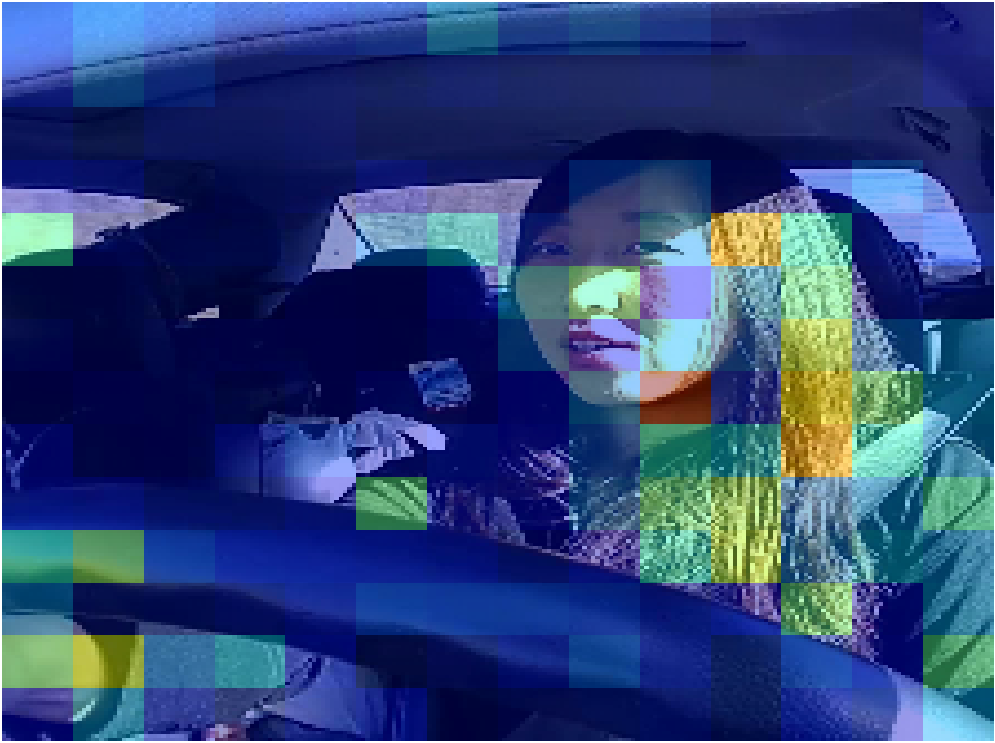}
\includegraphics[width=\sizeafigtwo\linewidth,height=\sizeafigtwo\linewidth]{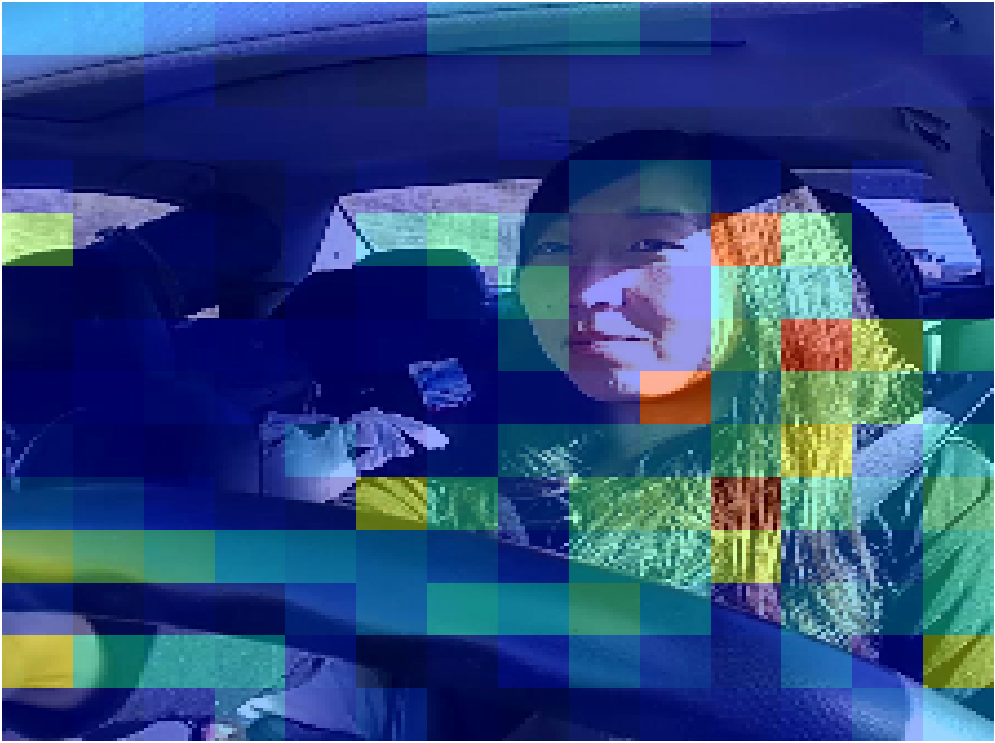}
\includegraphics[width=\sizeafigtwo\linewidth,height=\sizeafigtwo\linewidth]{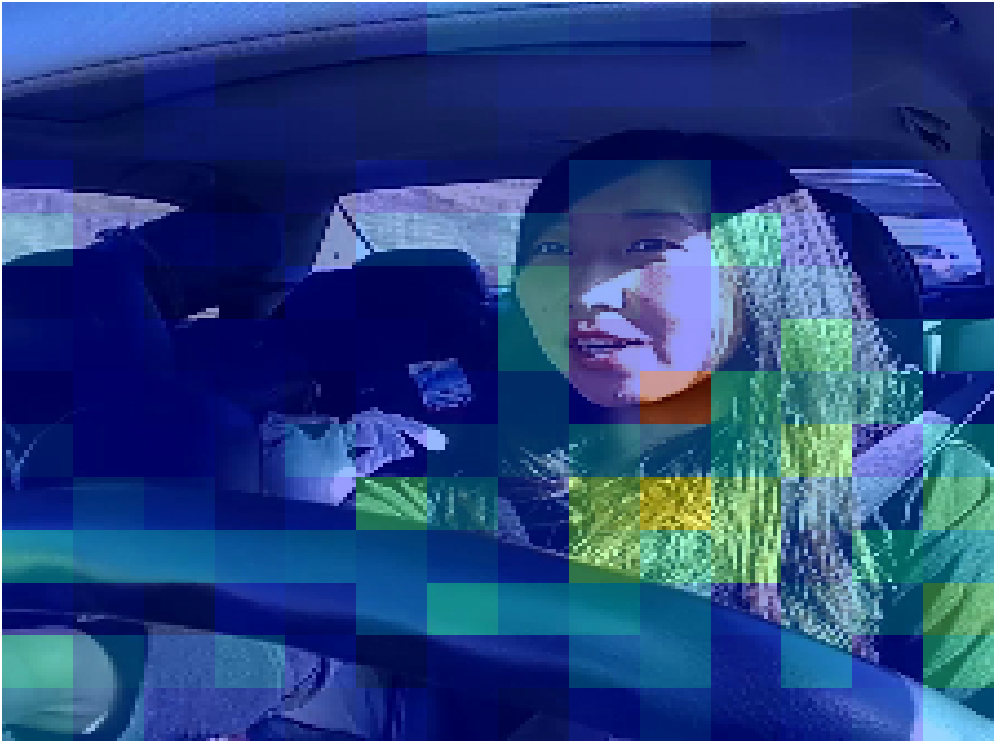}
\includegraphics[width=\sizeafigtwo\linewidth,height=\sizeafigtwo\linewidth]{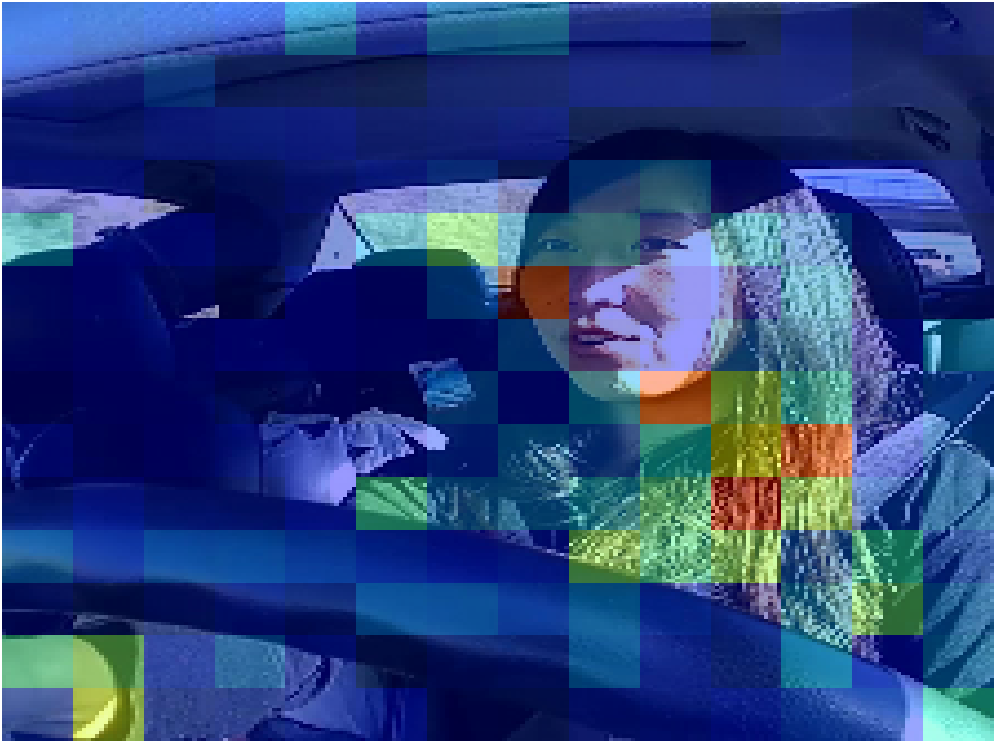}\\
Multi-View  with Episodic Memory
\vspacefigtwo
\end{minipage}
\begin{minipage}{\textwidth}
\centering
\fontsizefigtwo
\includegraphics[width=\sizeafigtwo\linewidth,height=\sizeafigtwo\linewidth]{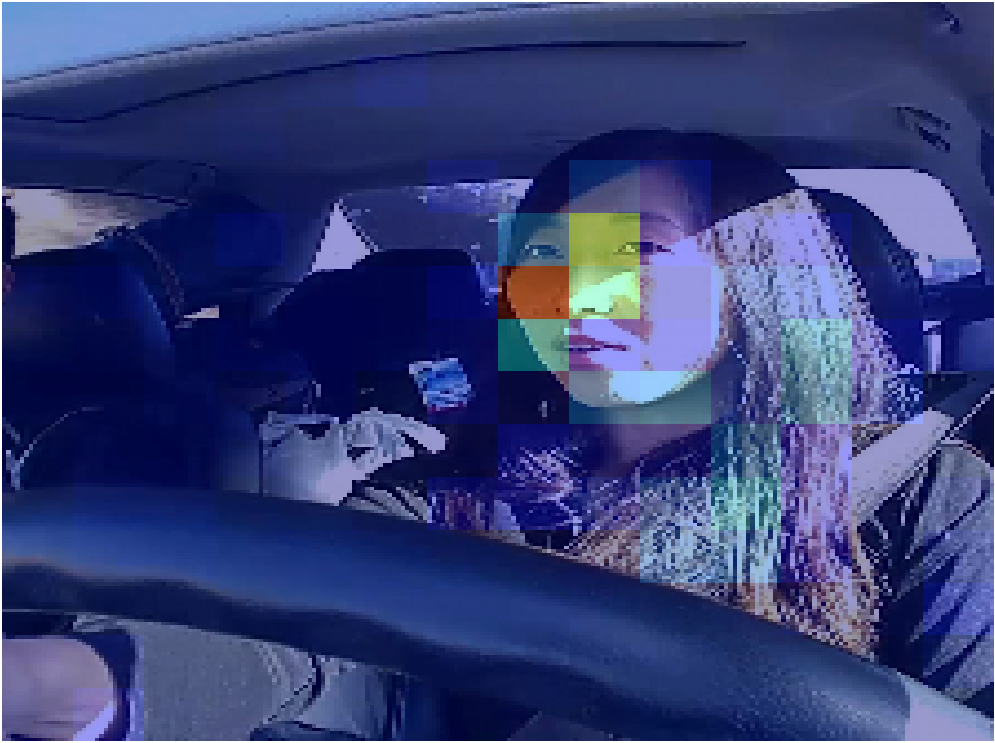}
\includegraphics[width=\sizeafigtwo\linewidth,height=\sizeafigtwo\linewidth]{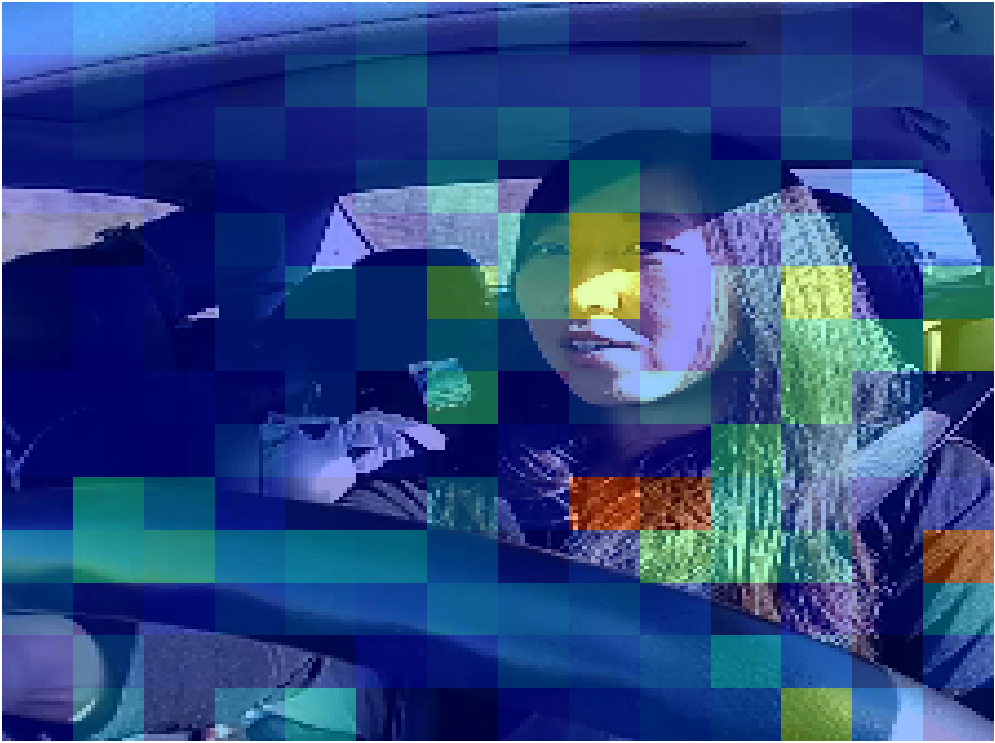}
\includegraphics[width=\sizeafigtwo\linewidth,height=\sizeafigtwo\linewidth]{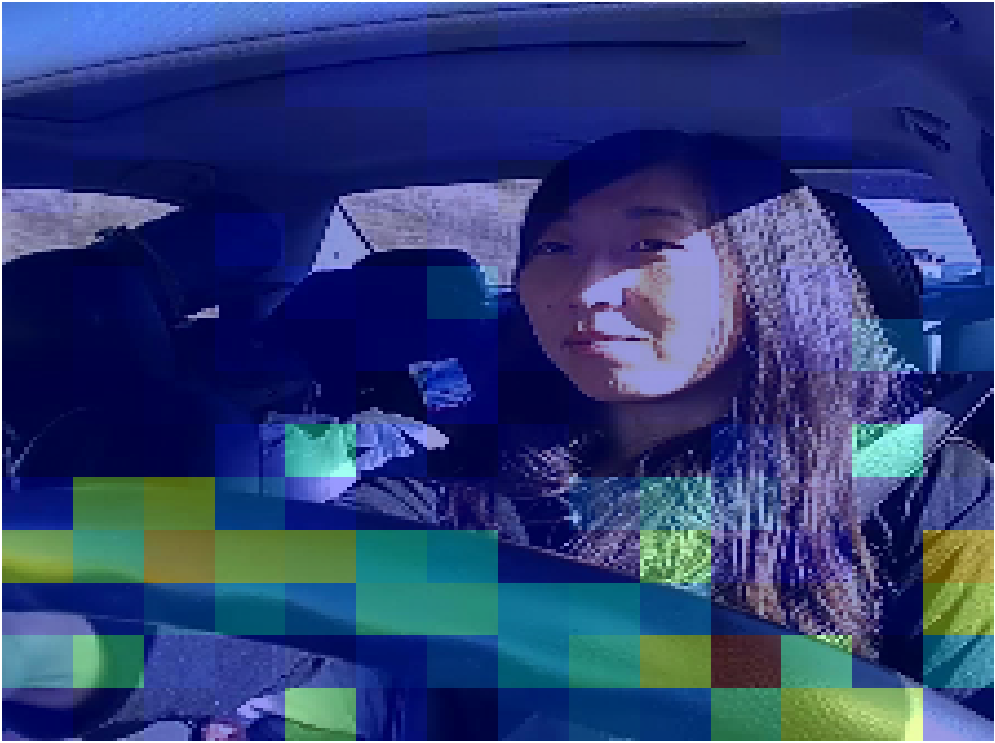}
\includegraphics[width=\sizeafigtwo\linewidth,height=\sizeafigtwo\linewidth]{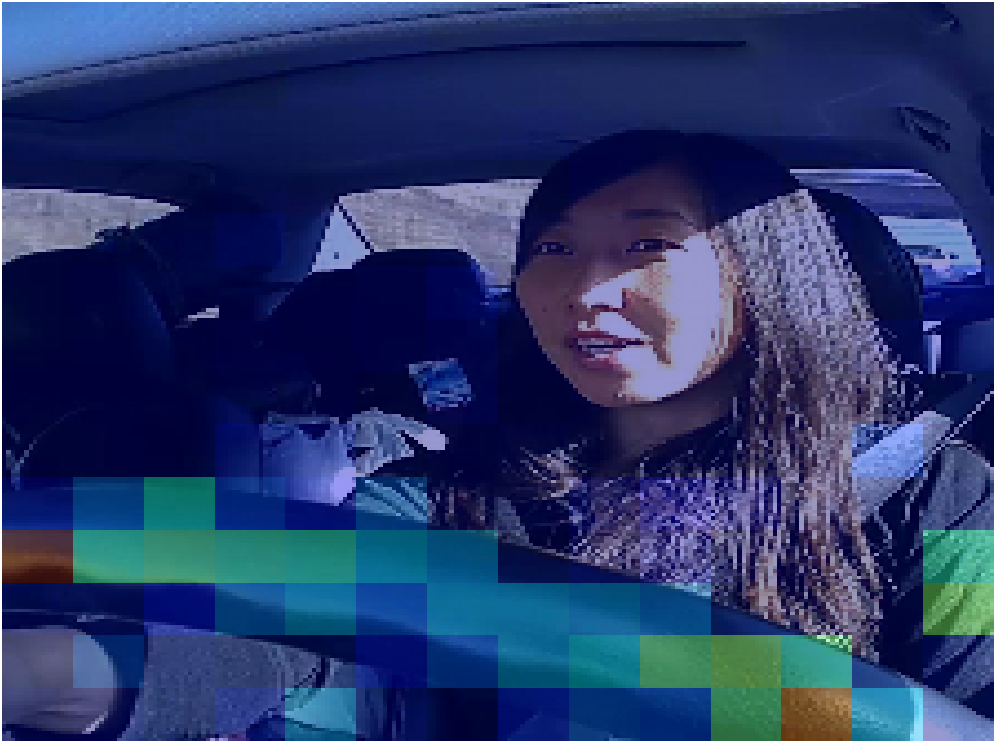}
\includegraphics[width=\sizeafigtwo\linewidth,height=\sizeafigtwo\linewidth]{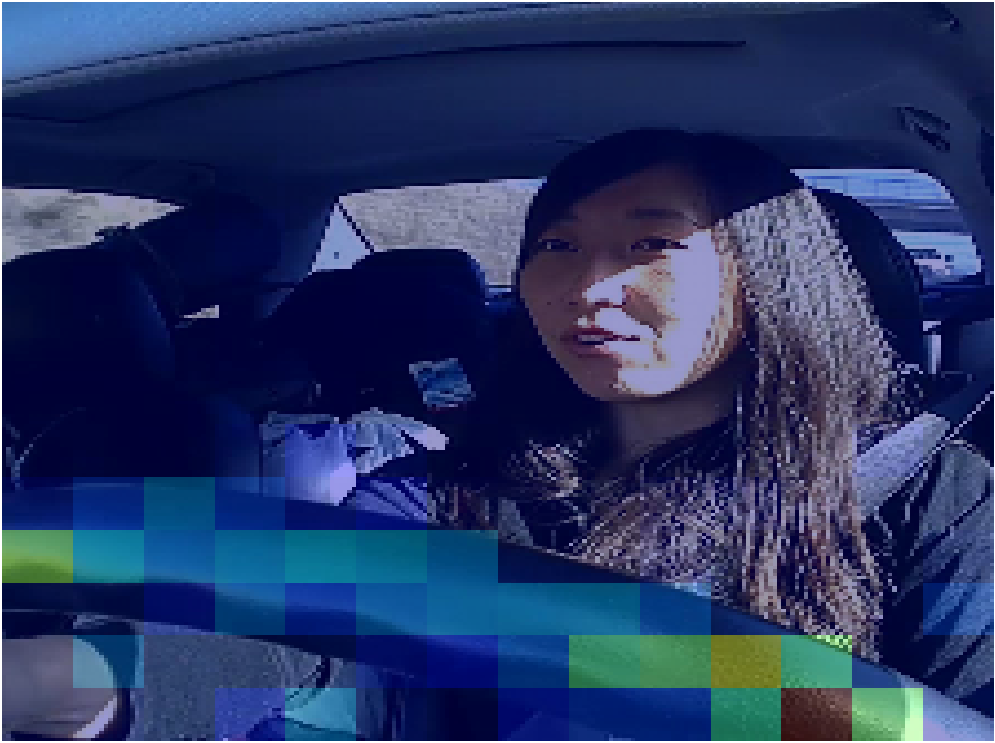}\\
Single-View with EM and CC
\vspacefigtwo
\end{minipage}
\begin{minipage}{\textwidth}
\hspace{0.3mm}
\begin{tikzpicture}[
    arrow/.style = {dashed,thick}]
\draw[arrow] (\recl,0) -- (\recl,\rech) -- (\recw,\rech) -- (\recw,0) -- cycle;
\draw[arrow] (\recl,0) -- (\recw,\rech);
\draw[arrow] (\recl,\rech) -- (\recw,0);
\end{tikzpicture}
\vspacefigtwo
\end{minipage}
\begin{minipage}{\textwidth}
\centering
\fontsizefigtwo
\includegraphics[width=\sizeafigtwo\linewidth,height=\sizeafigtwo\linewidth]{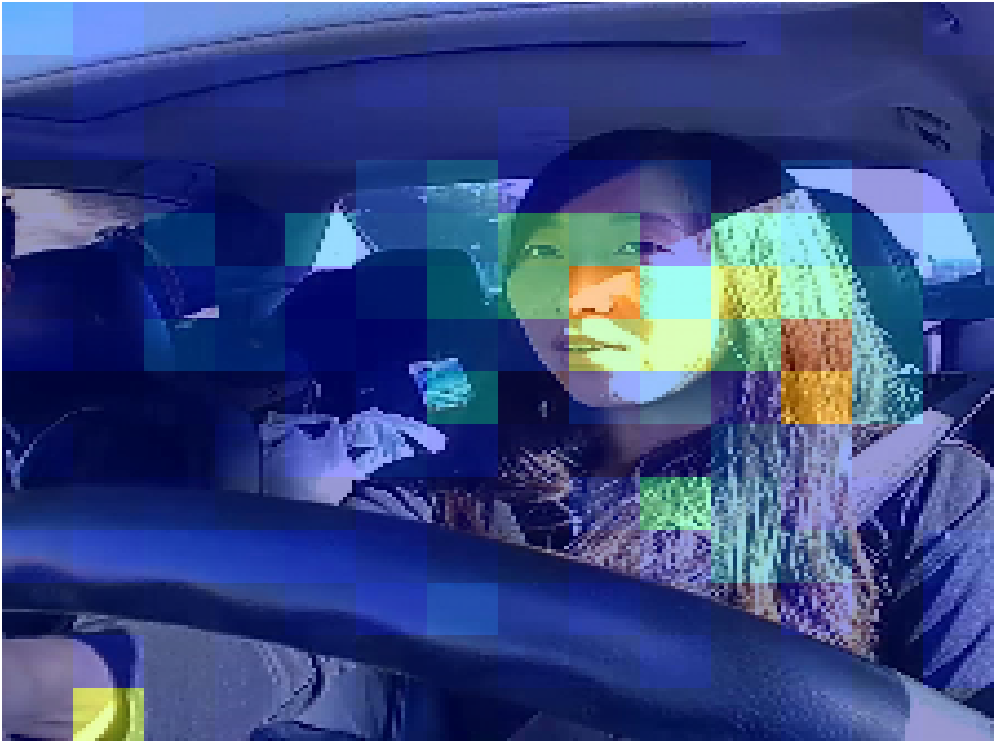} 
\includegraphics[width=\sizeafigtwo\linewidth,height=\sizeafigtwo\linewidth]{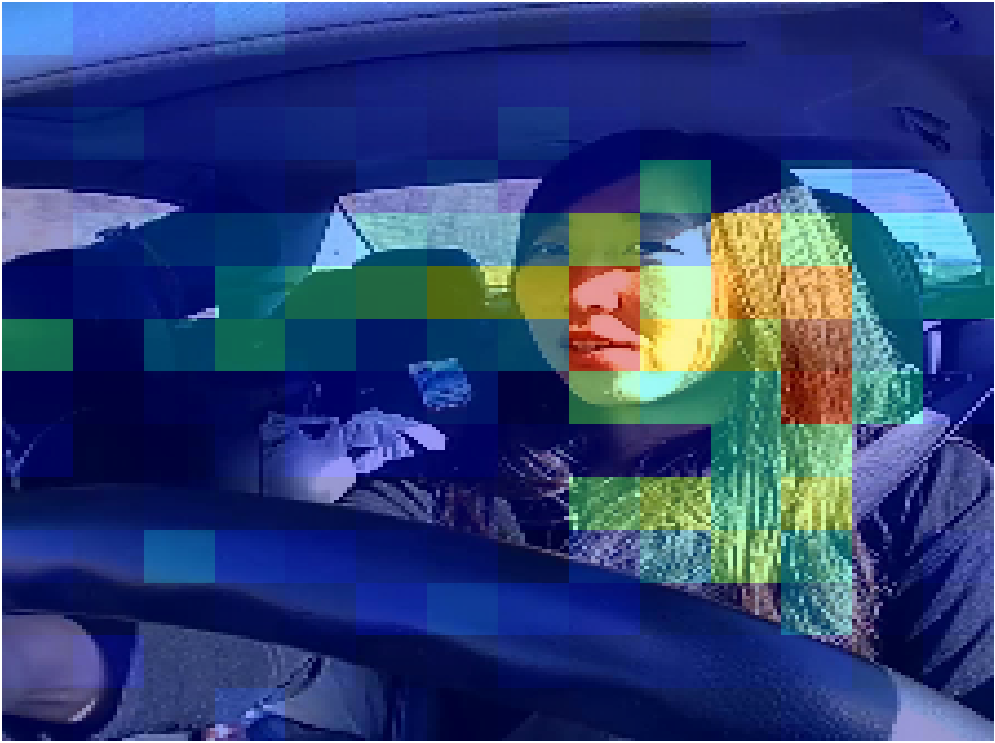}
\includegraphics[width=\sizeafigtwo\linewidth,height=\sizeafigtwo\linewidth]{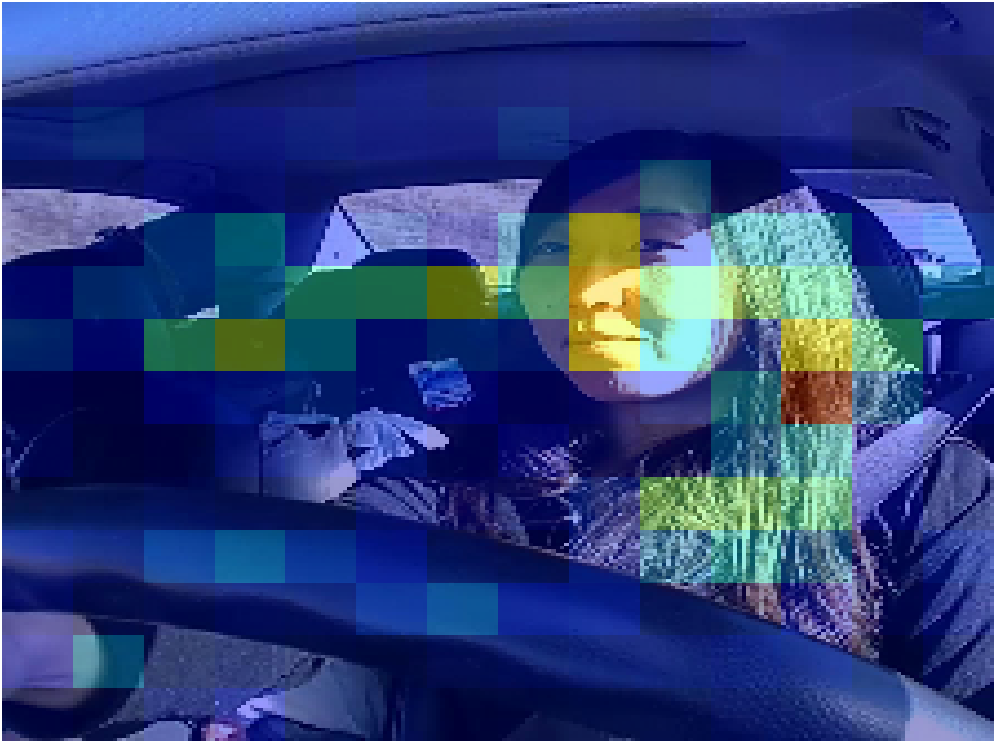}
\includegraphics[width=\sizeafigtwo\linewidth,height=\sizeafigtwo\linewidth]{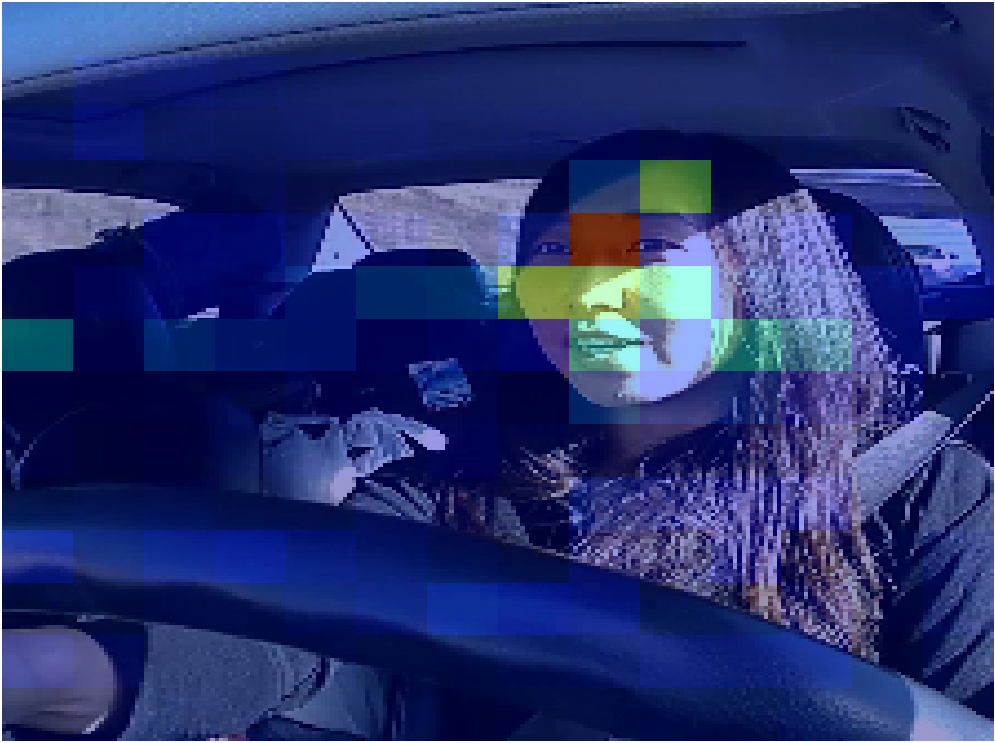}
\includegraphics[width=\sizeafigtwo\linewidth,height=\sizeafigtwo\linewidth]{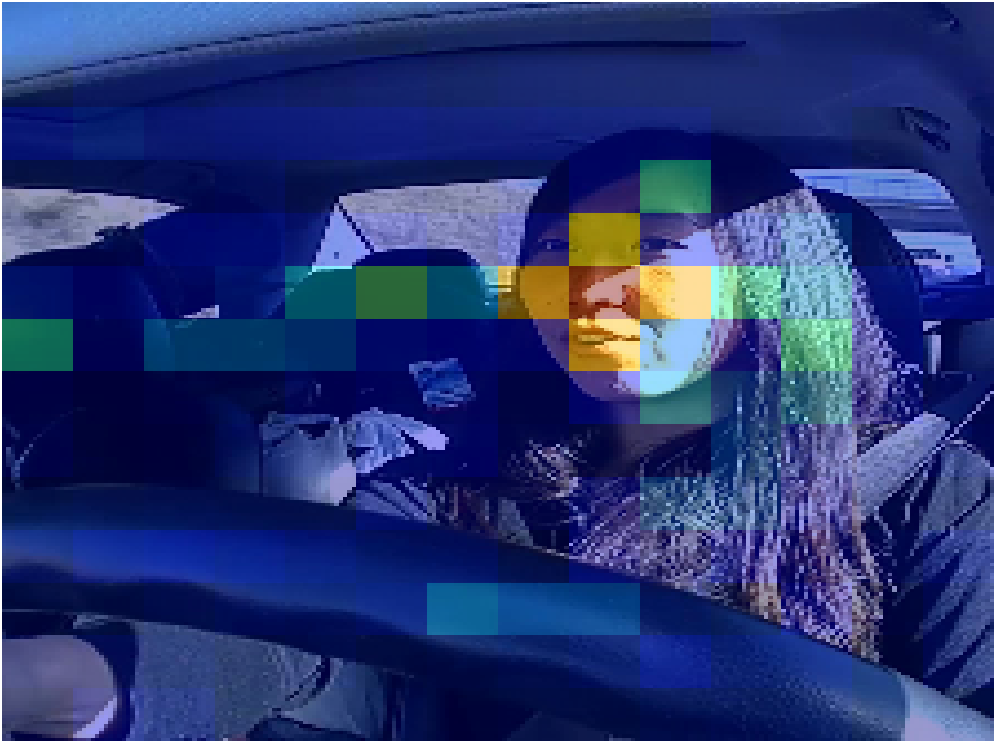}\\
Single-View with EM and CC
\end{minipage}
\end{minipage} 
\begin{minipage}{\sizerfigtwo\textwidth}
\centering
\begin{minipage}{\textwidth}
\begin{tikzpicture}[
    arrow/.style = {thick,-stealth}]
\node (A) at (0, 0) {\textbf{Time Step 1}};
\node (B) at (\lineend, 0) {\textbf{Time Step 5}};
\draw [arrow] (A) -- (B);
\end{tikzpicture}
\vspacefigtwo
\end{minipage}
\begin{minipage}{\textwidth}
\centering
\fontsizefigtwo
\includegraphics[width=\sizeafigtwo\linewidth,height=\sizeafigtwo\linewidth]{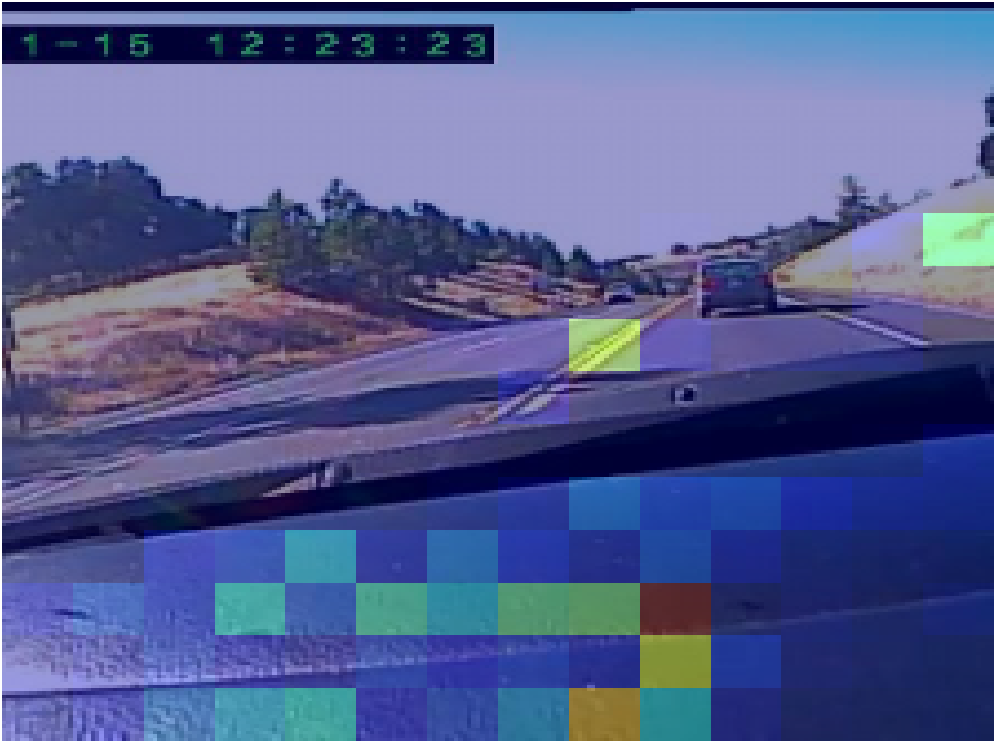} 
\includegraphics[width=\sizeafigtwo\linewidth,height=\sizeafigtwo\linewidth]{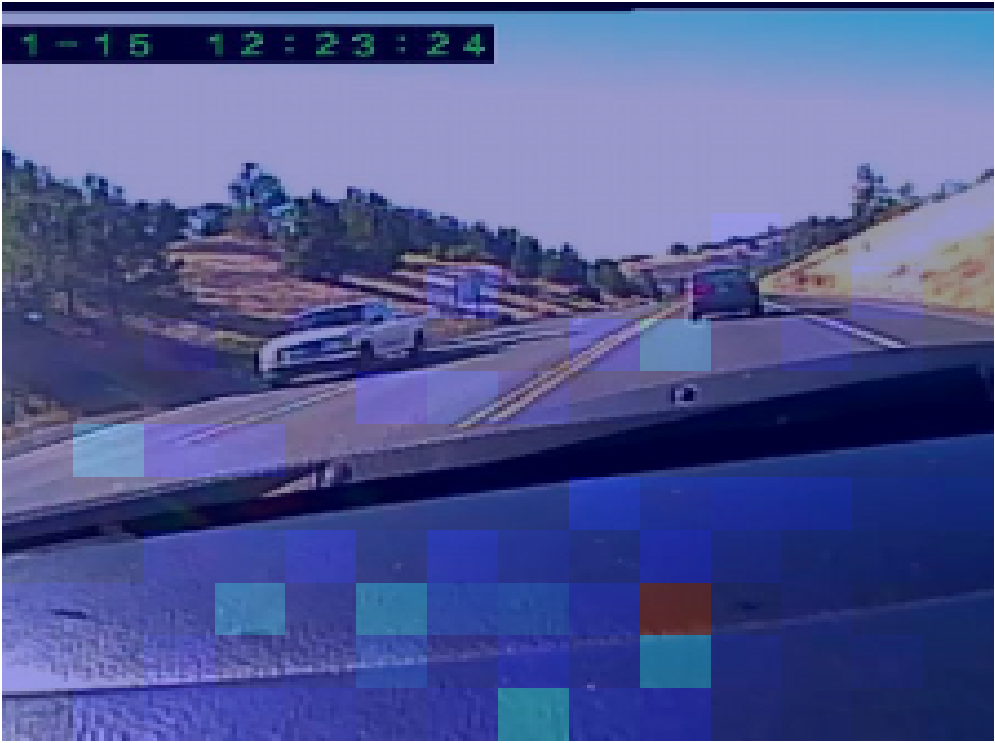}
\includegraphics[width=\sizeafigtwo\linewidth,height=\sizeafigtwo\linewidth]{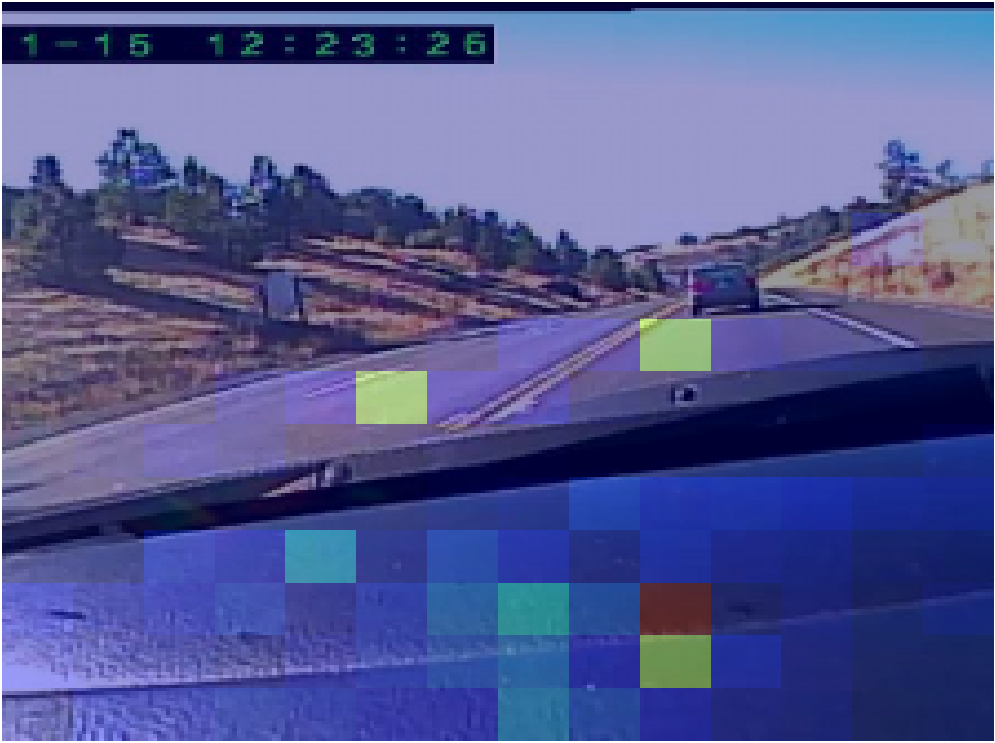}
\includegraphics[width=\sizeafigtwo\linewidth,height=\sizeafigtwo\linewidth]{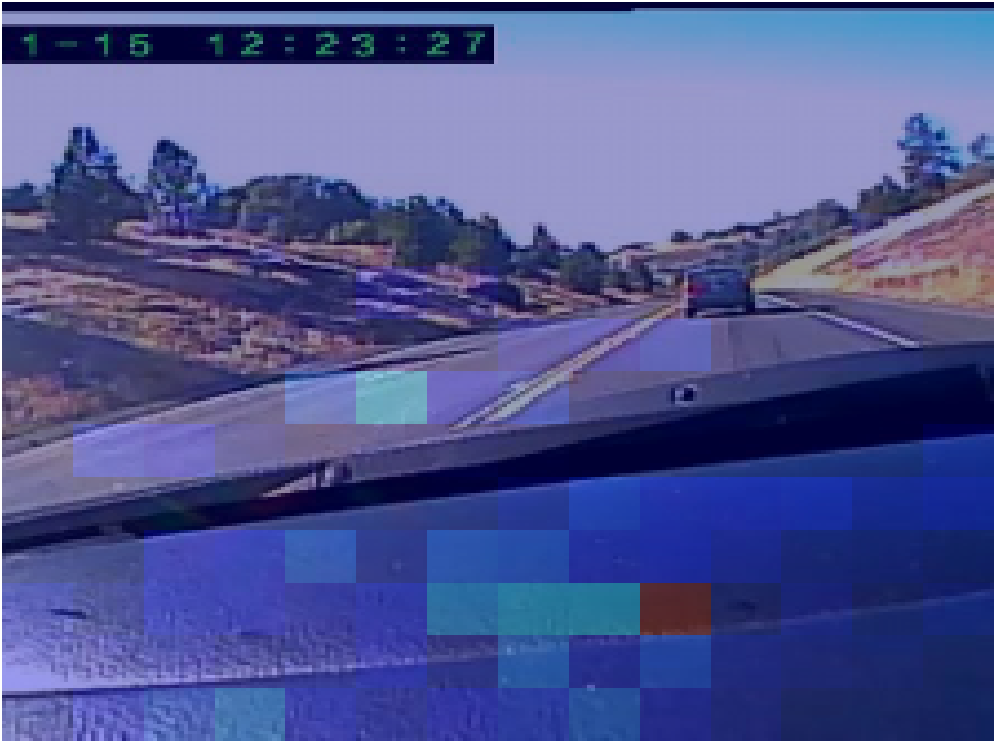}
\includegraphics[width=\sizeafigtwo\linewidth,height=\sizeafigtwo\linewidth]{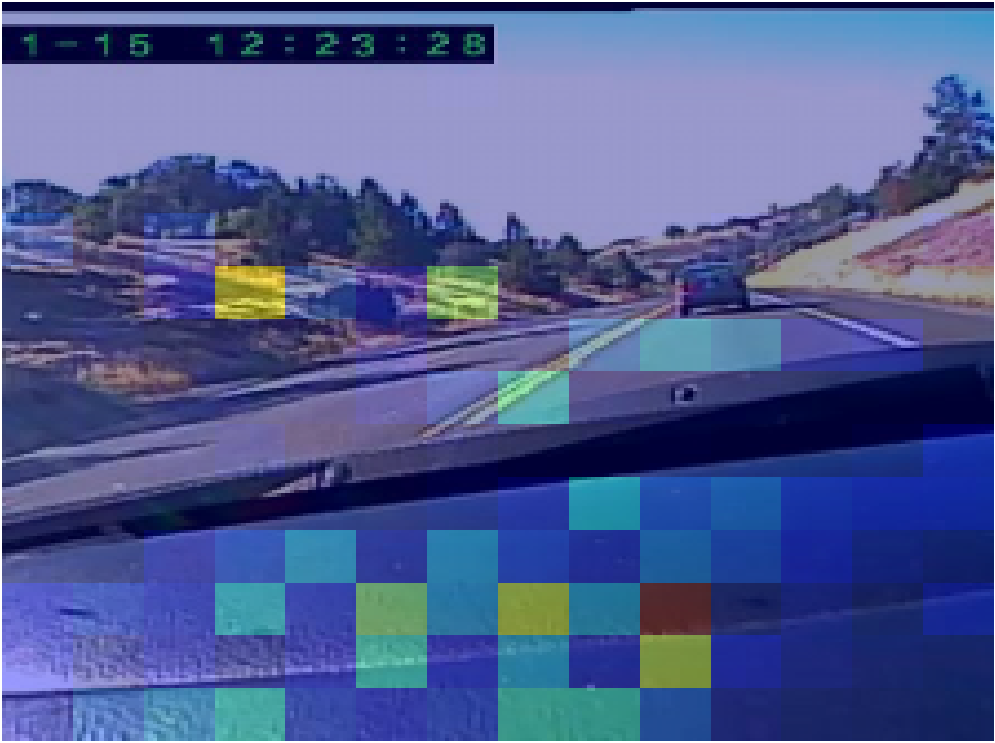}\\
Multi-View with Context Consistency
\vspacefigtwo
\end{minipage}
\begin{minipage}{\textwidth}
\centering
\fontsizefigtwo
\includegraphics[width=\sizeafigtwo\linewidth,height=\sizeafigtwo\linewidth]{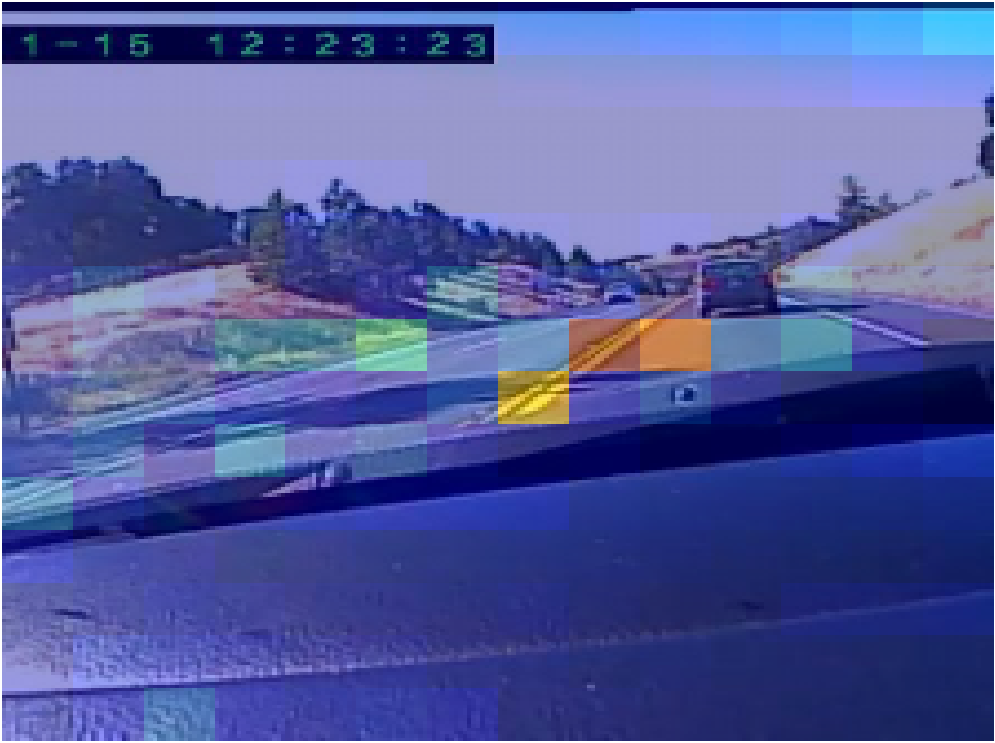} 
\includegraphics[width=\sizeafigtwo\linewidth,height=\sizeafigtwo\linewidth]{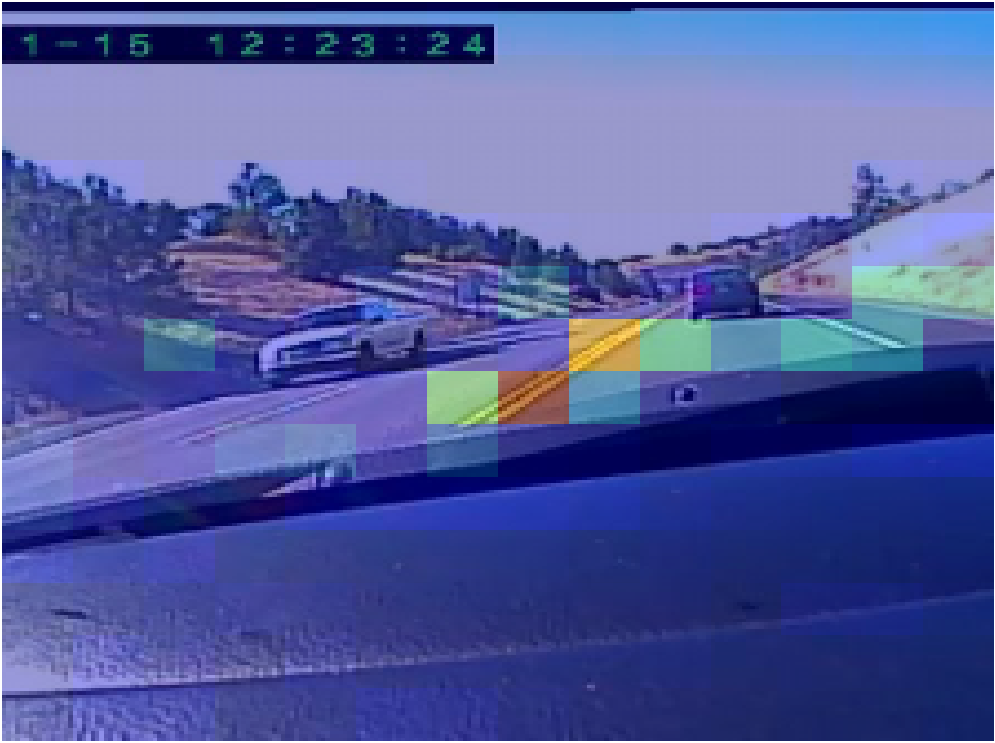}
\includegraphics[width=\sizeafigtwo\linewidth,height=\sizeafigtwo\linewidth]{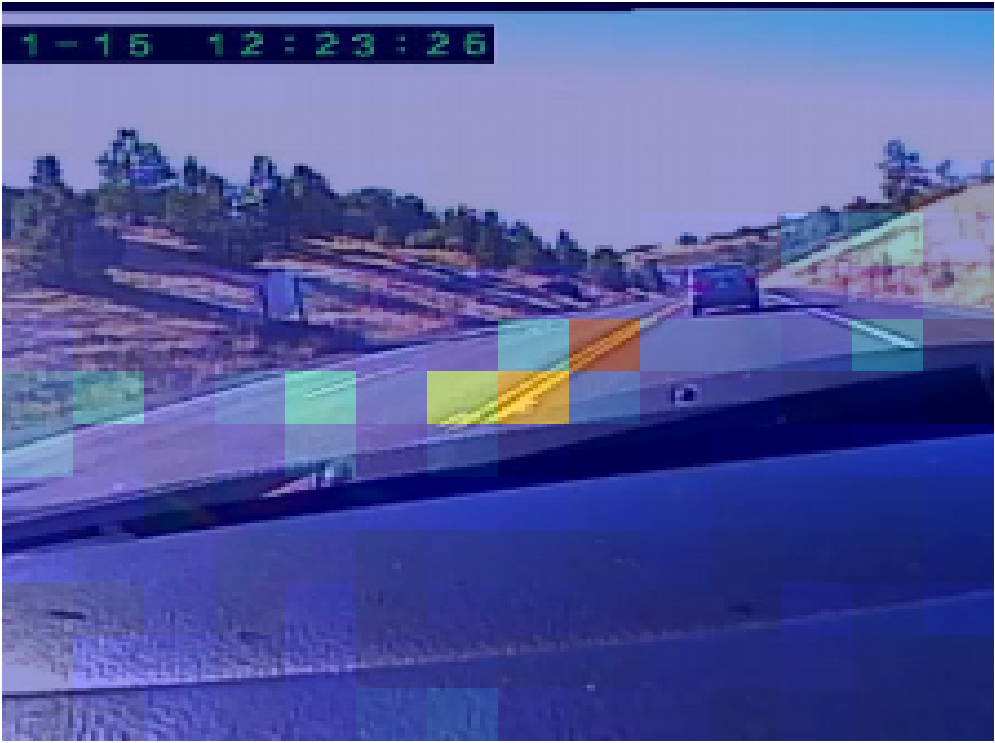}
\includegraphics[width=\sizeafigtwo\linewidth,height=\sizeafigtwo\linewidth]{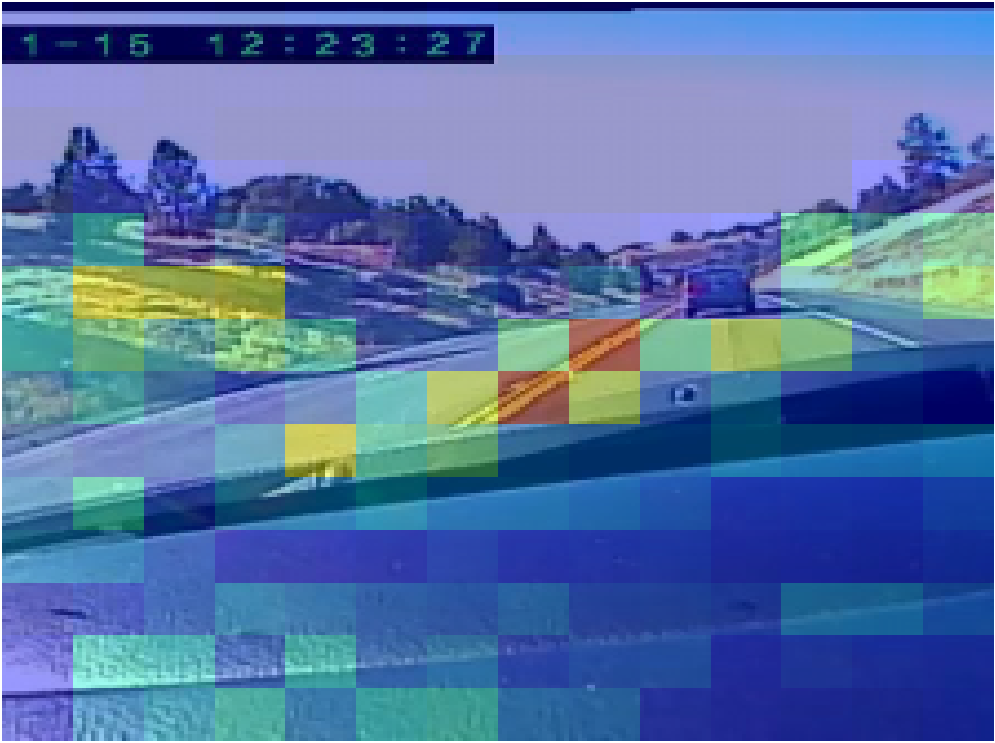}
\includegraphics[width=\sizeafigtwo\linewidth,height=\sizeafigtwo\linewidth]{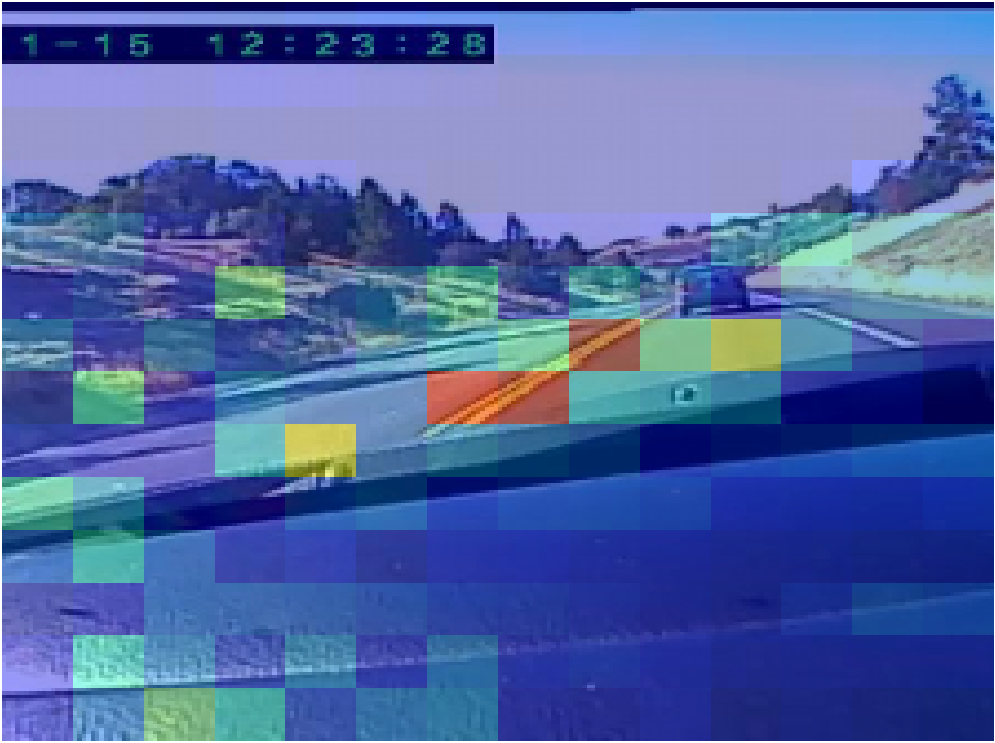}\\
Cross-View with Episodic Memory
\vspacefigtwo
\end{minipage}
\begin{minipage}{\textwidth}
\hspace{0.3mm}
\begin{tikzpicture}[
    arrow/.style = {dashed,thick}]
\draw[arrow] (\recl,0) -- (\recl,\rech) -- (\recw,\rech) -- (\recw,0) -- cycle;
\draw[arrow] (\recl,0) -- (\recw,\rech);
\draw[arrow] (\recl,\rech) -- (\recw,0);
\end{tikzpicture}
\vspacefigtwo
\end{minipage}
\begin{minipage}{\textwidth}
\centering
\fontsizefigtwo
\includegraphics[width=\sizeafigtwo\linewidth,height=\sizeafigtwo\linewidth]{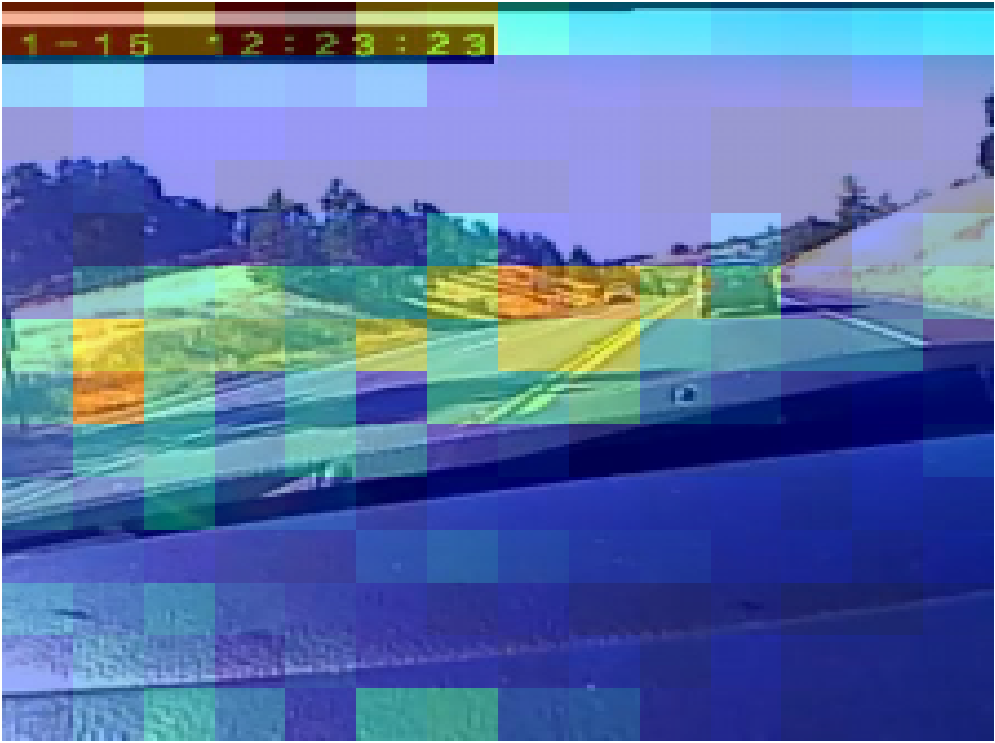} 
\includegraphics[width=\sizeafigtwo\linewidth,height=\sizeafigtwo\linewidth]{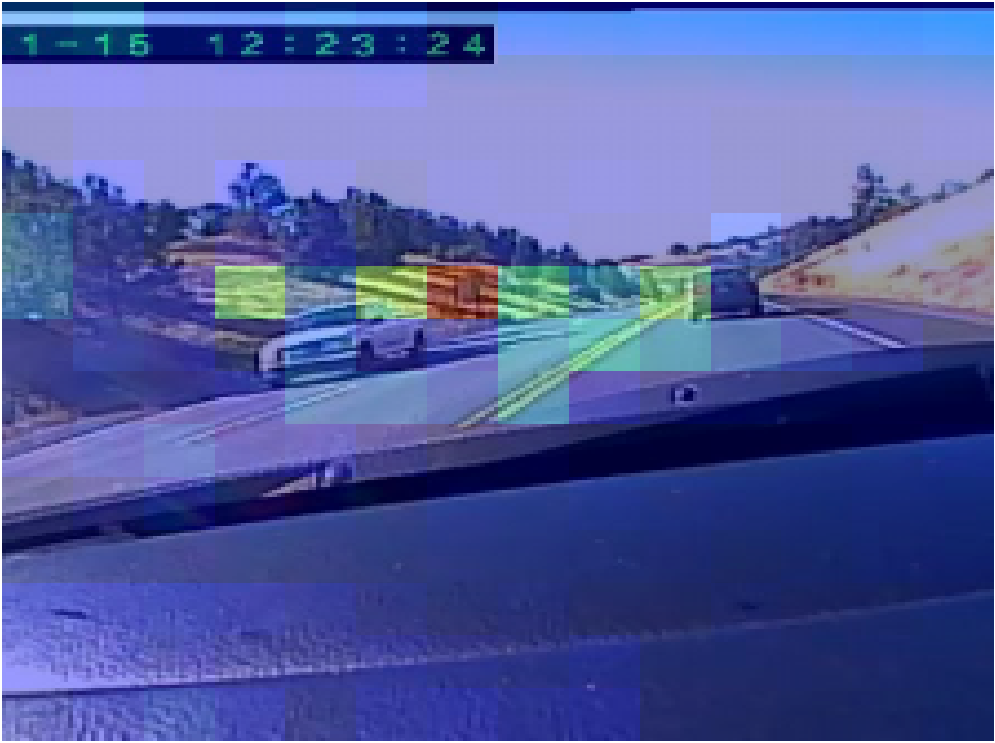}
\includegraphics[width=\sizeafigtwo\linewidth,height=\sizeafigtwo\linewidth]{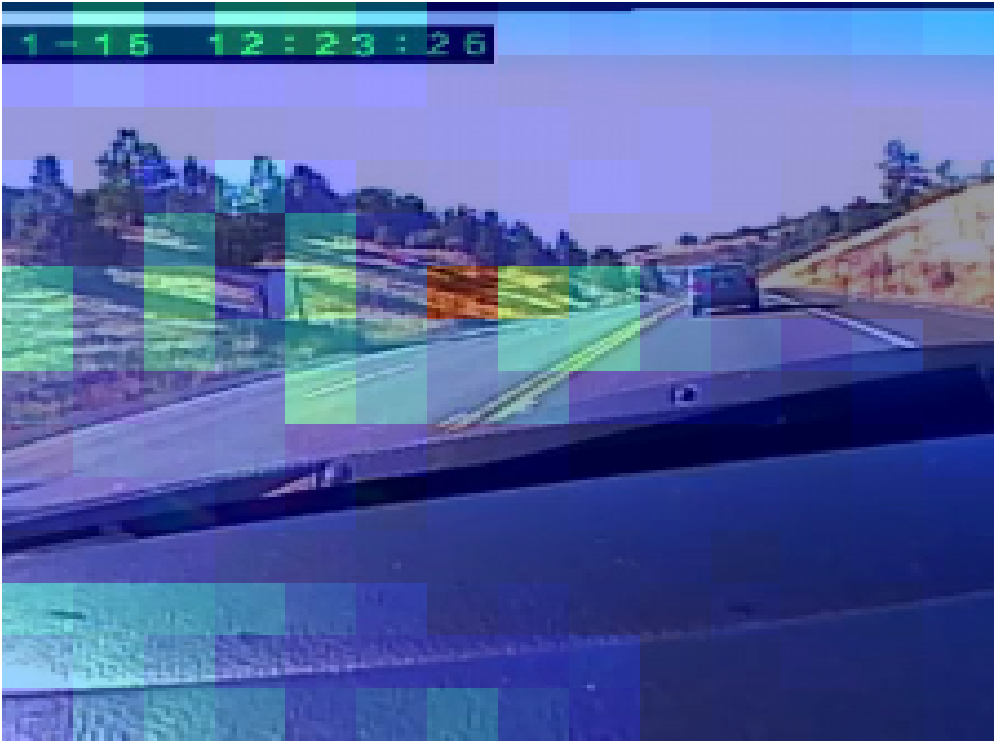}
\includegraphics[width=\sizeafigtwo\linewidth,height=\sizeafigtwo\linewidth]{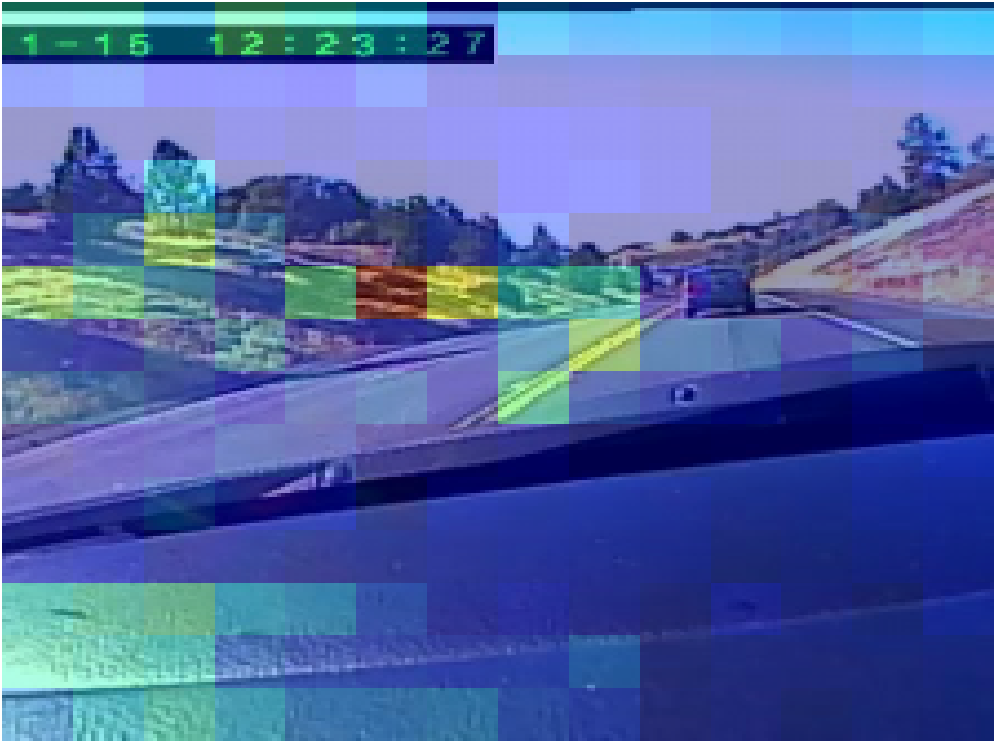}
\includegraphics[width=\sizeafigtwo\linewidth,height=\sizeafigtwo\linewidth]{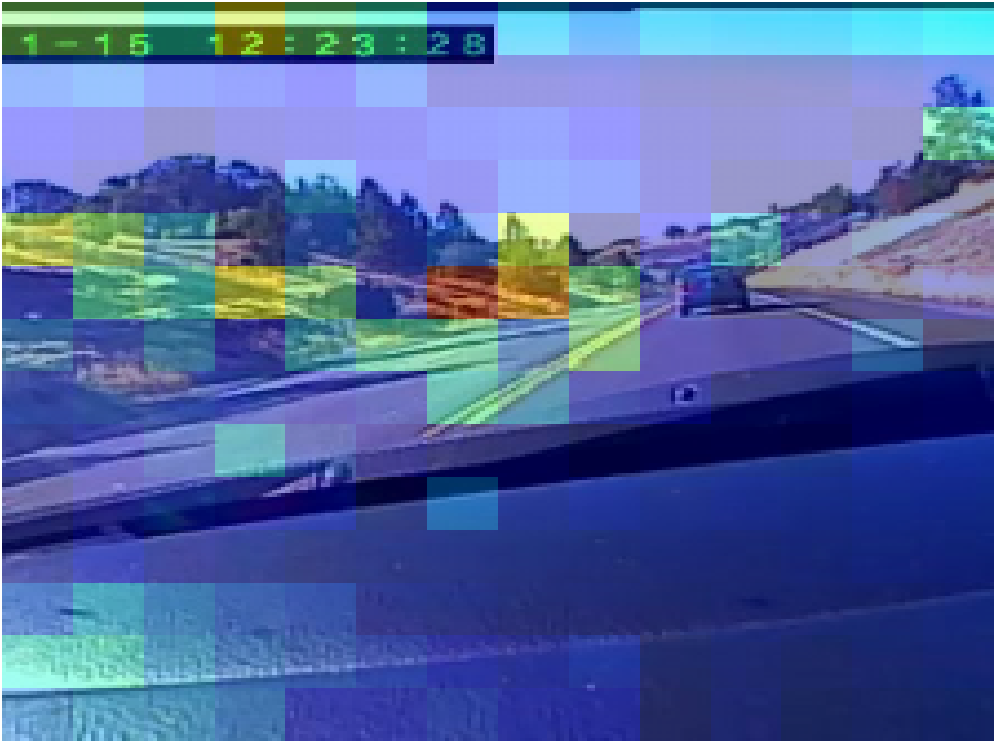}\\
Single-View with EM and CC
\vspacefigtwo
\end{minipage}
\begin{minipage}{\textwidth}
\centering
\fontsizefigtwo
\includegraphics[width=\sizeafigtwo\linewidth,height=\sizeafigtwo\linewidth]{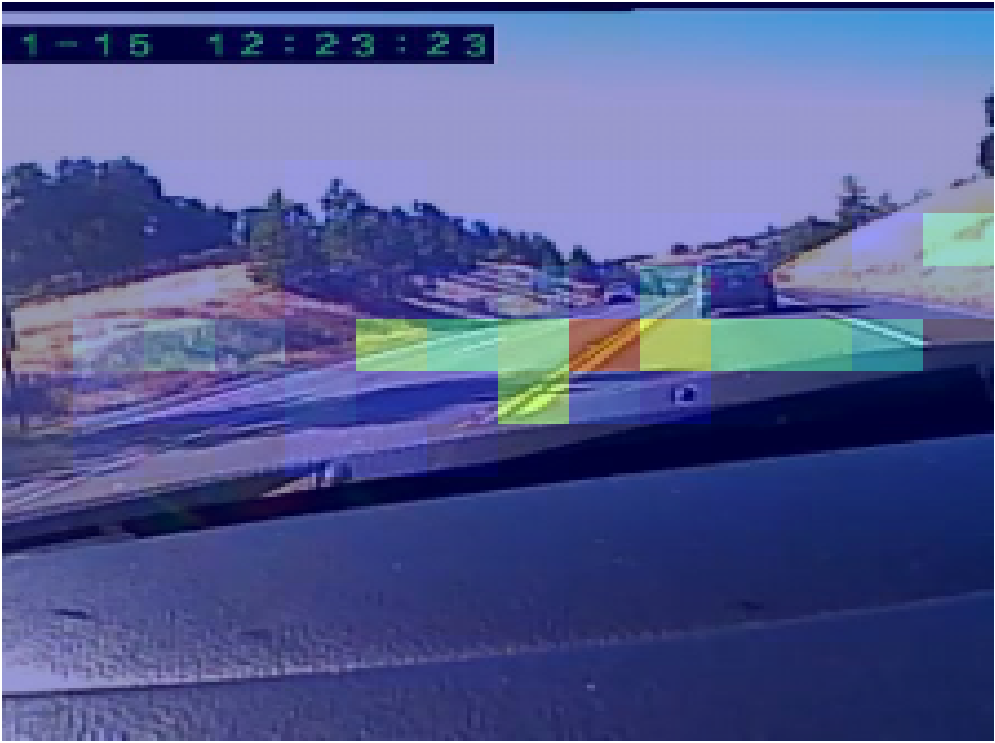} 
\includegraphics[width=\sizeafigtwo\linewidth,height=\sizeafigtwo\linewidth]{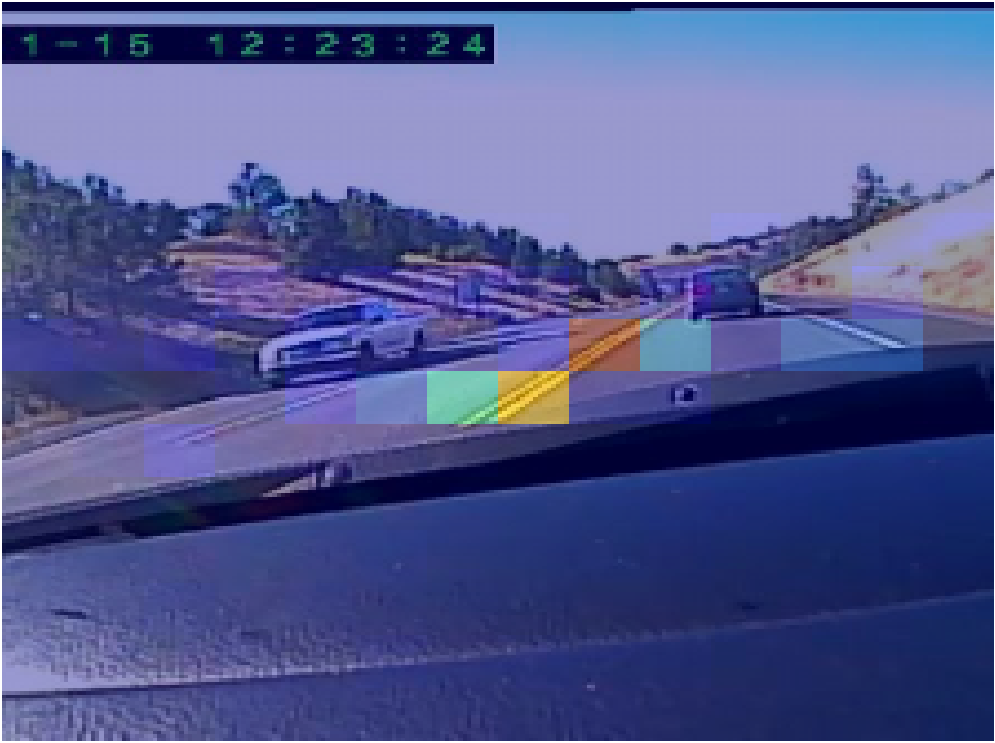}
\includegraphics[width=\sizeafigtwo\linewidth,height=\sizeafigtwo\linewidth]{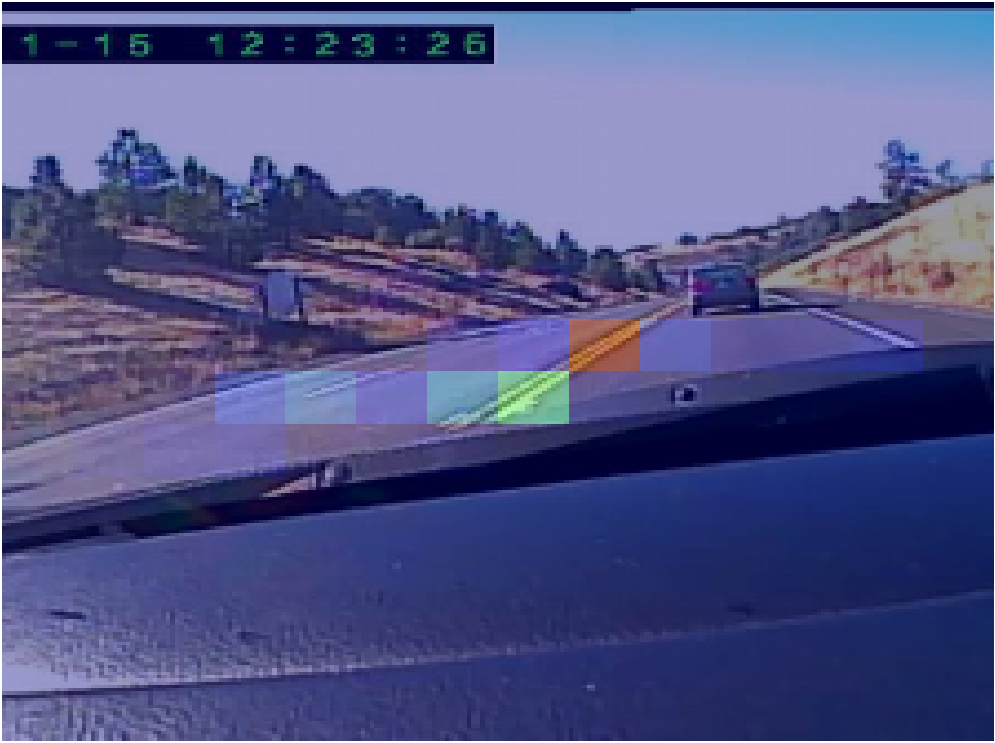}
\includegraphics[width=\sizeafigtwo\linewidth,height=\sizeafigtwo\linewidth]{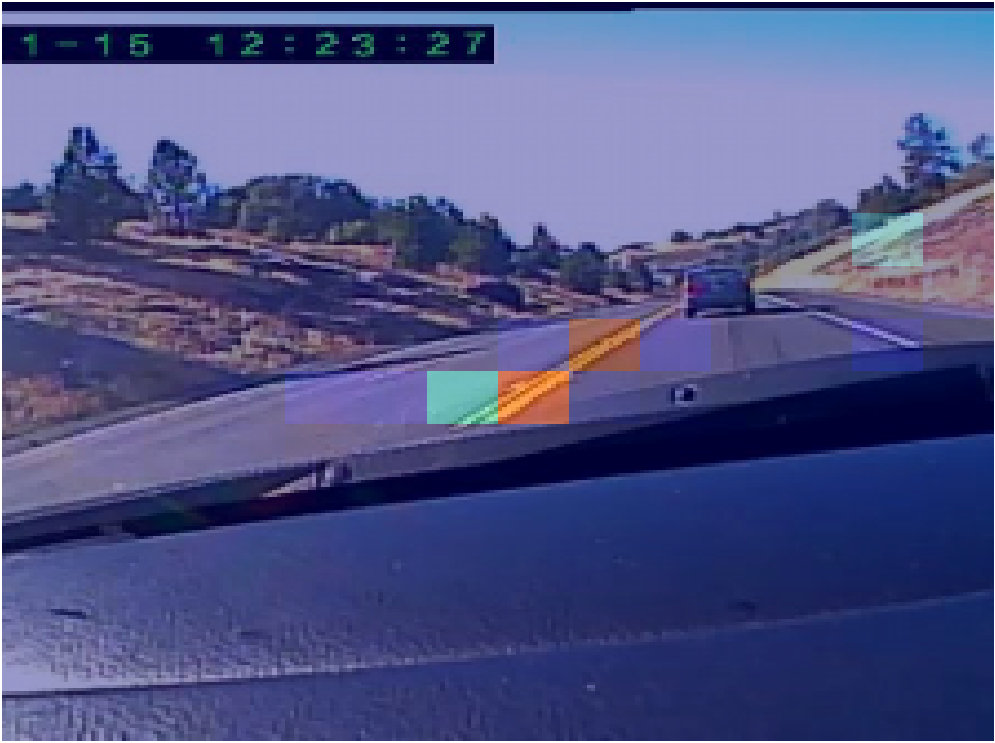}
\includegraphics[width=\sizeafigtwo\linewidth,height=\sizeafigtwo\linewidth]{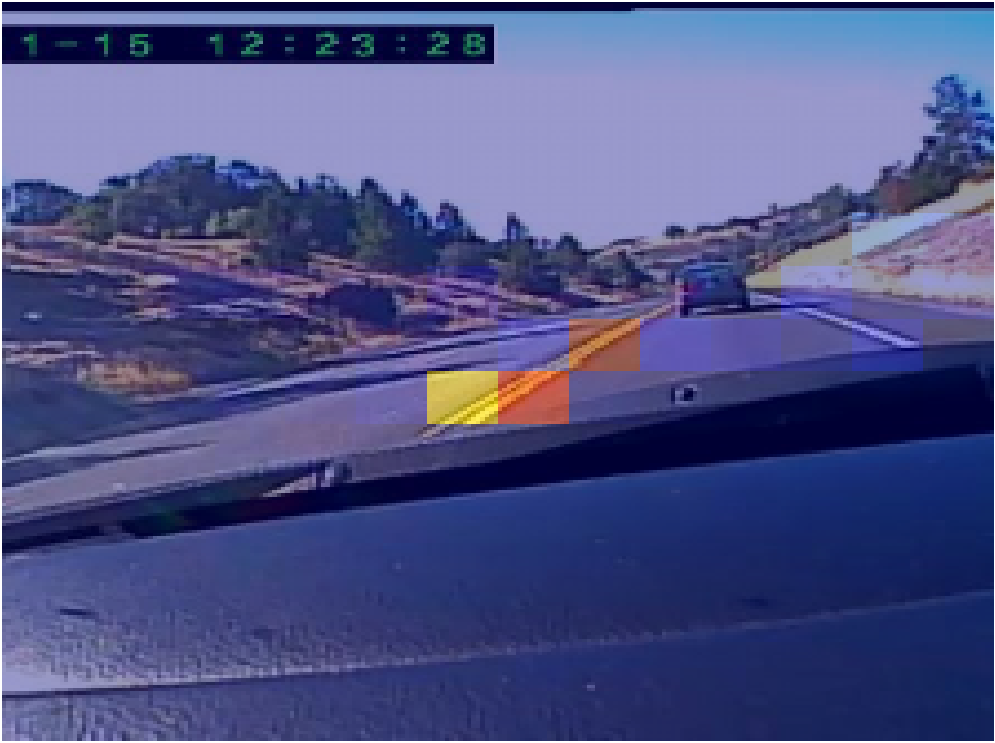}\\
Multi-View with EM and CC
\end{minipage}
\end{minipage} 
\end{minipage}

\caption{
Visualization of the attention maps generated by the proposed CEMFormer model, highlighting the influence of the episodic memory (EM), context consistency (CC), and multi-view input in identifying the most crucial regions of multi-camera streams. The attention maps are overlaid on the original images, with red indicating the highest level of focus on a region. The left and right frames showcase in-cabin and external views, respectively.
}
\label{fig:attn_map}
\end{figure*}

\subsection{Ablation Study}
\label{sec:ab}

\paragraph{Module Analysis}
To investigate the effect of different modules, we conduct ablation experiments to further verify the impact of the episodic memory (EM) and the context-consistency loss (CC) by modifying one component while keeping the other one fixed. Both in-cabin and external camera views are used in the experiments. According to the results in \cref{tab:ab_m}, the episodic memory mechanism contributes a 4.42\% improvement in F1 score compared to the baseline model. Meanwhile, the context-consistency loss results in a 5.44\% improvement in F1 score. The two components complement each other in terms of performance and variance, and when used together, they achieve an accuracy of 85.37\% and an F1 score of 87.09\%. We also visualize the contribution of each module in \cref{sec:qualitative_results}.



\paragraph{Influence of Episodic Memory Tokens}
We present the results of selecting the number of episodic memory tokens based on the comparison experiment in \cref{tab:abk}. The best performance in accuracy and F1 score is achieved when $K=4$. We also notice that larger values of $K$ lead to higher variance. The parameter $K$ is task-specific and is influenced by the number of input views and the complexity of the task.

\paragraph{Model Latency} We evaluate the latency of the proposed CEMFormer model with both single-view or double-view inputs. The latency is measured on an RTX 3090 Ti GPU. The results presented in ~\cref{tab:fps} demonstrates that the CEMFormer model achieves satisfactory real-time performance, with single-view and double-view inputs achieving 22.08 and 15.56 frames per second, respectively. This performance level indicates that the model's computational cost is suitable for real-time applications.

\begin{table}[!tb]
\centering
\caption{Real-time FPS performance of the CEMFormer model}
\begin{tabular}{l|c|c}
\toprule
Data Source & Parameters & Frames per second (FPS)\\
\midrule
Single-view & 86.6M & 22.08\\
Dual-view & 87.3M & 15.56\\
\bottomrule
\end{tabular}
\label{tab:fps}
\end{table}

\subsection{Qualitative Results \& Interpretability}
\label{sec:qualitative_results}
To offer a thorough understanding of the CEMFormer model's performance during online inference, we present detailed visualizations that demonstrate how the individual transformer modules contribute to identifying crucial regions within multi-view streams. The well-trained CEMFormer model performs remarkably in most scenarios, as illustrated in~\cref{fig:attn_map} using one episode as an example.

We visualize attention maps for both in-cabin and external views in the last layer of the spatial-temporal encoder. Attention scores are employed to generate attention maps, which are then superimposed onto the original images. These attention scores are reshaped according to their original spatial positions to produce the attention maps, with red representing the highest level of focus on a specific region. Images displayed in the same row show results from the same model. The following observations can be made based on the visualization results:

\paragraph{Episodic Memory} Comparing the first and last rows, it becomes evident that when episodic memory is unavailable, image frames are encoded independently, preventing the model from fusing historical information. In contrast, applying episodic memory allows the model to fuse historical data, gradually reducing uncertainty as the temporal context increases. This observation holds for both in-cabin and external views.

\paragraph{Context Consistency} Comparing the second and last rows, we can see that without context consistency, the attention map appears more divergent, indicating suboptimal performance. This observation supports our claim that the context consistency loss functions as a regularizer, assisting the model in reducing uncertainty and concentrating on the most important regions. This consistency is observed in both in-cabin and external views.

\paragraph{Multi-View} When comparing single-view attention maps to those of multiple views, the latter exhibits superior outcomes. In the in-cabin view, the single-view model's focus is inconsistent, initially concentrating on the driver's face before shifting to the steering wheel. In contrast, the cross-view results prioritize the driver's face correctly. Likewise, in the external view, the single-view results pay more attention to off-road regions, while the cross-view results focus on the road—particularly the center line.

\section{Related Work}
\paragraph{Video Understanding Models} Video understanding models are designed to enable computers to comprehend and interpret video content in a manner similar to that of humans. These models process videos as sequences of images and employ various techniques, such as handcrafted features~\cite{laptev_learning_2008,peng_action_2014}, recurrent neural networks~\cite{sun_lattice_2017,li_recurrent_2018}, convolutional neural networks\cite{feichtenhofer_slowfast_2019,zhao_end--end_2020}, and transformer-based architectures~\cite{xu_long_2021,li_mvitv2_2022}, to extract and analyze the spatio-temporal features from videos. They then utilize this data to make predictions about the video content.

\paragraph{Assistive Features for Vehicles}
Modern vehicles are equipped with cameras and other sensors that continuously monitor the surrounding environment. These sensors, using multi-sensory fusion, provide various assistive features such as lane keeping, forward collision avoidance, and adaptive cruise control. These features not only warn drivers of potentially hazardous maneuvers but also enhance the overall driving experience. While driver monitoring for distraction and drowsiness has been extensively researched~\cite{rezaei_look_2014,ahmed_intelligent_2022,ma_vit-dd_2023}, our work focuses on building next-generation ADAS capable of anticipating maneuvers before they occur~\cite{liu_vision-cloud_2022}. This capability will not only improve current ADAS and driver monitoring techniques but also significantly enhance driver safety.

\section{Discussion and Conclusion}

In this work, we have proposed CEMFormer, a framework designed to predict driver intentions using in-cabin and external camera inputs. The model efficiently aggregates spatial-temporal information and employs the novel context-consistency loss to incorporate driving context as an auxiliary supervision signal during training. Despite these advancements, there are some limitations and future directions worth exploring.

The design of the episodic memory module is based on the assumption that the most informative contextual cues appear shortly (typically less than 5 seconds) before the maneuver~\cite{morris_lane_2011}. This assumption may not always be accurate, which could limit the model's predictive capabilities in certain scenarios.

We observed a modest accuracy improvement when incorporating external data into in-cabin camera data. One possible explanation for this limited improvement could be that when both in-cabin and external cameras are available, the prediction relies predominantly on the in-cabin data, which provides information about the driver's behavior, such as head and eye movements. The external view offers supplementary traffic information, which might not be directly related to predicting driver intentions. As a result, the advantage of combining the two data streams may be diminished by the noise introduced by irrelevant information. Alternatively, it is possible that processing the forward-facing camera data is more challenging than the in-cabin camera data due to the dynamic nature of the traffic environment.

Moreover, we demonstrated that incorporating traffic context information in the form of a finite set of traffic context encodings can enhance driver intention prediction performance. However, leveraging traffic navigation data collected from High Definition (HD) maps could potentially provide even greater benefits, as it delivers centimeter-level offline location services for ADAS and minimizes environmental interference with real-time streaming camera data~\cite{tang_thma_2022}. Consequently, future work could involve incorporating additional sensors or integrating with existing HD map systems to further improve the proposed CEMFormer model.


In conclusion, our work contributes to ongoing efforts to improve traffic safety and paves the way for further advancements in driver intention prediction and personalized driving assistance systems.

\section*{Acknowledgment}
This work is funded by the Digital Twin Roadmap of InfoTech Labs, Toyota Motor North America. The contents of this paper only reflect the views of the authors, who are responsible for the facts and the accuracy of the data presented herein. The contents do not necessarily reflect the official views of Toyota Motor North America.









\bibliographystyle{IEEEtran}
\bibliography{my}

\end{document}